%% file: main.tex
\documentclass{article}

\PassOptionsToPackage{numbers, compress}{natbib}

\usepackage[main, final]{neurips_2025}

\usepackage[utf8]{inputenc} 
\usepackage[T1]{fontenc}    
\usepackage{hyperref}       
\usepackage{url}            
\usepackage{booktabs}       
\usepackage{amsfonts}       
\usepackage{nicefrac}       
\usepackage{microtype}      
\usepackage{xcolor}         
\usepackage{graphicx}
\usepackage{amsmath}
\usepackage{amssymb}
\usepackage{booktabs}
\usepackage{multirow}
\usepackage{nicematrix}
\usepackage{caption}

\title{One-Step Diffusion-Based Image Compression with Semantic Distillation}

\author{%
  Naifu Xue$^{1*}$, Zhaoyang Jia$^{2*}$, Jiahao Li$^3$, Bin Li$^3$, Yuan Zhang$^1$, Yan Lu$^3$ \\
  $^1$ Communication University of China \quad
  $^2$ University of Science and Technology of China\\
  $^3$ Microsoft Research Asia\\
  {\tt\small \{aaronxuenf, yzhang\}@cuc.edu.cn, \{jzy$\_$ustc\}@mail.ustc.edu.cn }\\
  {\tt\small \{li.jiahao, libin, yanlu\}@microsoft.com} \\
}

\begin{document}

\maketitle

\begin{abstract}

While recent diffusion-based generative image codecs have shown impressive performance, their iterative sampling process introduces unpleasant latency.
In this work, we revisit the design of a diffusion-based codec and argue that multi-step sampling is not necessary for generative compression.
Based on this insight, we propose OneDC, a \textbf{One}-step \textbf{D}iffusion-based generative image \textbf{C}odec—that integrates a latent compression module with a one-step diffusion generator.
Recognizing the critical role of semantic guidance in one-step diffusion, we propose using the hyperprior as a semantic signal, overcoming the limitations of text prompts in representing complex visual content.
To further enhance the semantic capability of the hyperprior, we introduce a semantic distillation mechanism that transfers knowledge from a pretrained generative tokenizer to the hyperprior codec.
Additionally, we adopt a hybrid pixel- and latent-domain optimization to jointly enhance both reconstruction fidelity and perceptual realism.
Extensive experiments demonstrate that OneDC achieves SOTA perceptual quality even with one-step generation, offering over \textbf{39\%} bitrate reduction and \textbf{20$\times$} faster decoding compared to prior multi-step diffusion-based codecs. 
Project: \url{https://onedc-codec.github.io/}

\end{abstract}

\begingroup
\makeatletter
\renewcommand\thefootnote{}

\def\@makefntext#1{\noindent\footnotesize #1}


\footnotetext{%
$^{*}$Naifu Xue and Zhaoyang Jia are visiting students at Microsoft Research Asia.
\vspace{-4mm}}
\makeatother
\endgroup

\input{section_1}
\input{section_2}

\input{section_3}
\input{section_4}

\bibliographystyle{plainnat}
\bibliography{references}

\appendix

\clearpage
\input{section_6_append}

\clearpage

\input{section_7_append_figures}
\clearpage
\input{section_8_absolute_number}

\end{document}

%% file: section_1.tex
\section{Introduction}

The rapid growth in image data has led to increased storage and transmission costs, heightening the need for efficient, high-quality image compression techniques.
While VAE-based learned image compression (LIC) \cite{liu2023learned, zhao2023universal} has surpassed traditional codecs (e.g., VVC \cite{bross2021overview}) in rate-distortion (RD) performance, it often produces blurry details at low bitrates \cite{mentzer2020high} since they are typically optimized for objective distortion rather than human perception \cite{blau2019rethinking}.
To mitigate this, some approaches \cite{mentzer2020high, muckley2023improving} focus on optimizing visual quality by incorporating perceptual metrics and adversarial losses to enable generative image compression. 
Despite these advances, such generative models tend to introduce artifacts under extreme compression, degrading the realism of reconstructed images.

Recently, diffusion-based generative codecs \cite{careil2023towards, li2024towards} have been introduced to enhance reconstruction quality by leveraging the powerful content synthesis capabilities of pretrained models. 
While these methods significantly improve perceptual realism, they may occasionally generate content that deviates from the original input, thereby compromising reconstruction fidelity. 
In addition, their inherently iterative sampling process leads to substantial computational overhead, making them notably slower than conventional VAE-based codecs.

In standard diffusion-based image generation, the model progressively refines a noisy signal through iterative denoising, beginning with coarse structures and gradually synthesizing high-frequency details \cite{yang2023diffusion}.
However, when applied to image compression, the task changes: at low bitrates, codecs typically retain coarse structures in the compressed latents. As a result, the decoder is primarily responsible for plausibly reconstructing high-frequency content from the preserved low-frequency information.
This observation motivates our central hypothesis: \textit{given the compressed latent, multi-step sampling is not necessary for decoding, and a carefully designed one-step alternative could suffice.}
Although recent advances in one-step diffusion \cite{song2023consistency, yin2024improved, kang2024distilling} offer promising tools for accelerating inference, their potential remains underexplored in the context of image compression.

\begin{figure}[t]
    \centering
    \includegraphics[width=1.0\linewidth]{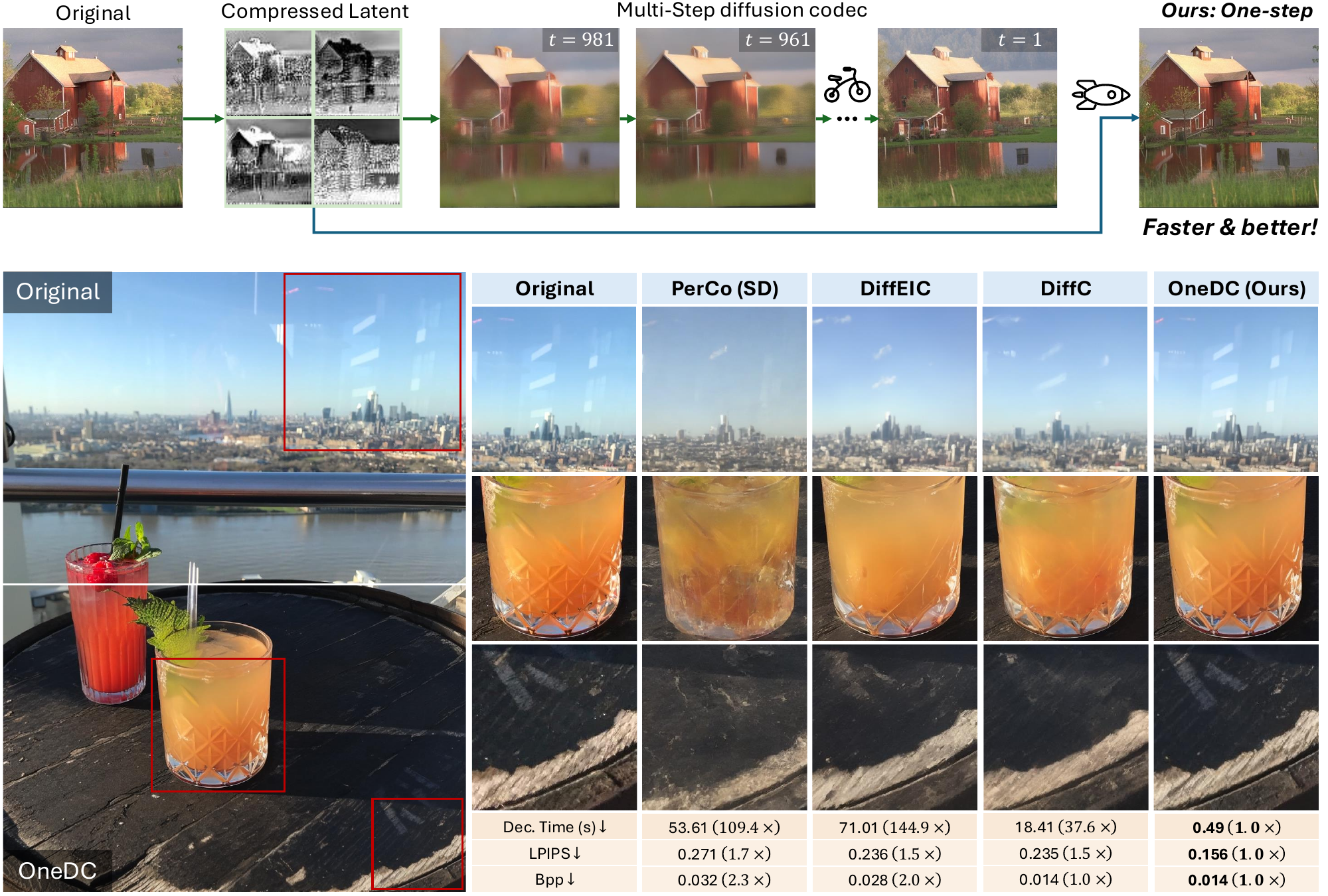}
    \vspace{-4mm}
    \caption{\textit{Top}: multi-step sampling is not essential for image compression; intermediate results are from DiffEIC~\cite{li2024towards}. \textit{Bottom}: Visual comparisons including existing open-sourced multi-step diffusion codecs~\cite{careil2023towards, li2024towards, vonderfecht2025lossy} and our proposed \textbf{one-step} codec. Our method achieves the highest visual quality at the lowest bitrate while offering significantly faster decoding.}
    \label{fig:1}
    \vspace{-5mm}
\end{figure}

In this paper, we introduce \textbf{OneDC} (\textbf{One}-step \textbf{D}iffusion-based generative \textbf{C}odec), a novel framework for ultra-low bitrate image coding. 
OneDC integrates a latent compression module with a one-step diffusion generator: the former encodes the image into compact latents, while the latter synthesizes high-frequency details conditioned on the latent. 
Since the pixel-domain training is insufficient for guaranteeing perceptual quality at low bitrates \cite{jia2024generative}, we adopt a hybrid-domain training strategy. 
Specifically, we combine a pixel-domain perceptual loss to promote fidelity with a latent-domain diffusion distillation objective \cite{yin2024improved} to enhance realism. 
This design allows OneDC to effectively leverage the pretrained diffusion model while achieving a balanced trade-off among compression ratio, perceptual quality, and decoding efficiency, as illustrated in Fig.~\ref{fig:1}.

Moreover, we further explore the role of semantic guidance (i.e., the input of the cross-attention layers \cite{rombach2021highresolution}) within our framework. In one-step diffusion, such guidance is essential to compensate for the absence of multi-step refinement (see ablation in Section \ref{sec:4.3.1}). 
While existing diffusion models have utilized text prompts for conditioning, we argue that text is suboptimal for image compression due to two limitations: (1) natural language struggles to capture fine-grained or localized visual semantics, and (2) generating high-quality captions typically requires large-scale vision-language models (e.g., the large BLIP2 \cite{li2023blip} used in \cite{careil2023towards}), introducing substantial computational overhead. 
This raises a central question: \textit{Can we design more effective semantic representations than text for guiding one-step diffusion-based codecs—without incurring excessive computational cost?}

Recent studies have shown that the hyperprior in VAE-based codecs can capture high-level semantic information \cite{jia2024generative, qi2025generative}. Compared to textual prompts, hyperpriors provide more precise and spatially aligned semantic cues, making them well-suited for representing localized semantics in high-resolution images (see Fig.\ref{fig:2}). 
This insight motivates us to further enhance the representational capability of the hyperprior. 
Inspired by generative tokenizers \cite{esser2021taming}, where codebooks encode rich semantic content, we propose hyperprior semantic distillation—a training strategy that transfers semantic knowledge from a pretrained tokenizer to the hyperprior through a code prediction module \cite{zhou2022towards}. 
As illustrated in Fig.\ref{fig:2}, this approach results in reconstructions that are more semantically accurate and visually coherent.

Experiments show that OneDC achieves state-of-the-art (SOTA) performance in generative image compression, delivering over \textbf{39\%} bitrate reduction and \textbf{20$\times$} faster decoding compared to existing multi-step diffusion codecs.
Ablation studies further validate the effectiveness of our optimization strategy and the semantic distillation method for hyperprior.
Our contributions are as follows:

\begin{figure}[t]
    \centering
    \includegraphics[width=1.0\linewidth]{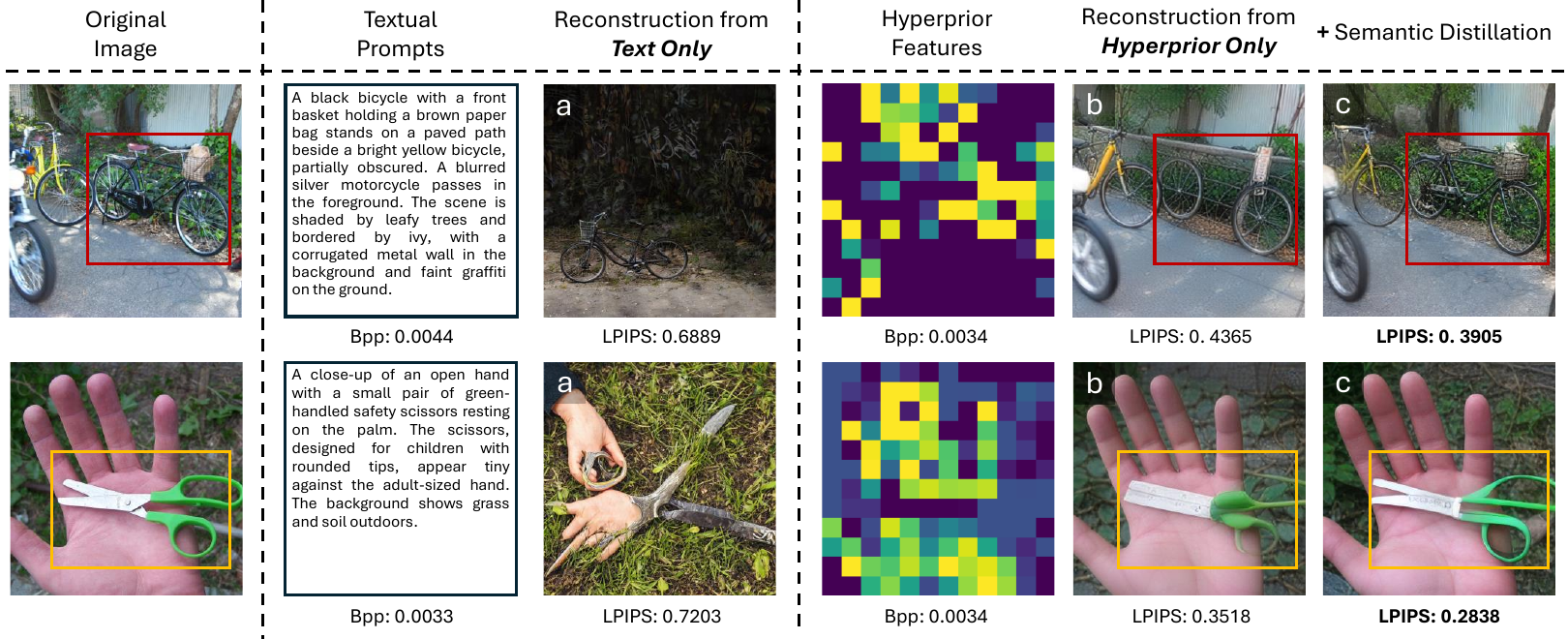}
    \vspace{-5mm}
    \caption{Reconstructions from different semantic guidance. (a) Text prompts (from GPT-4o~\cite{openai2024gpt4o}) struggle to capture complex visual semantics, resulting in severe distortions when using a pretrained text-to-image one-step diffusion model~\cite{yin2024improved}. (b) We finetune the model \cite{yin2024improved} for hyperprior-to-image generation. Hyperprior guidance yields more faithful reconstructions. (c) Our proposed semantic distillation further improves object-level accuracy, particularly in the highlighted regions.}
    \label{fig:2}
    \vspace{-5mm}
\end{figure}

\begin{itemize}
    \item We propose OneDC, a one-step diffusion codec comprising a latent compression module for compact feature encoding and a one-step diffusion generator for fast, high-quality decoding. A hybrid-domain training strategy further enhances both fidelity and perceptual realism.
    \item We identify the importance of high-level semantic guidance in one-step diffusion and highlight the limitations of text embeddings. To improve reconstruction quality, we introduce hyperprior features as an alternative and enhance them through semantic distillation.
    \item Extensive experiments show that OneDC achieves SOTA compression performance while offering significantly faster decoding than existing diffusion-based codecs, demonstrating the potential of one-step diffusion in generative compression.
\end{itemize}

%% file: section_2.tex
\section{Related Work}

\textbf{VAE-based Learned Image Compression.} 
Learned image compression has made rapid progress in RD performance. 
Ballé et al.~\cite{balle2017end} introduced an end-to-end framework with analysis/synthesis transforms and a factorized entropy model.
The hyperprior model~\cite{balle2018variational} enhanced entropy modeling by encoding side information (hyperprior) to predict distribution parameters, substantially improving RD performance.
Subsequent work combined hyperpriors with spatial context~\cite{minnen2018joint} and adopts more expressive distribution models such as Gaussian mixtures~\cite{cheng2020learned}.
Recent studies further refined transform architectures for stronger nonlinear representation~\cite{liu2023learned} and developed advanced entropy models for more compact symbol coding~\cite{jiang2023mlic}, together surpassing the VVC-Intra baseline~\cite{bross2021overview}.
Nevertheless, these models are typically optimized with distortion-oriented objectives such as MSE or MS-SSIM~\cite{wang2003multiscale}, which often over-smooth textures and degrade perceptual quality at low bitrates~\cite{mentzer2020high}, even with SOTA transforms and entropy models~\cite{liu2023learned,zhao2023universal,wang2023EVC}.

\textbf{VAE-based Generative Image Compression.} 
To improve perceptual quality at low bitrates, GAN-based approaches such as Agustsson et al.~\cite{Agustsson_2019_ICCV} and HiFiC~\cite{mentzer2020high} train VAE codecs to synthesize high-frequency details using perceptual losses~\cite{johnson2016perceptual} and adversarial discriminators.
MS-ILLM \cite{muckley2023improving} improves realism through a local likelihood model, while EGIC \cite{korber2024egic} introduces semantic-aware feedback via segmentation-conditioned discriminators. 
Other methods enhance compression by integrating generative tokenizers: Mao et al. \cite{mao2024extreme} utilize VQGAN \cite{esser2021taming} for compression, Xue et al. \cite{xue2024unifying} unify token generation and entropy coding, and Jia et al. \cite{jia2024generative} propose GLC that improves token compression via transform coding. 
More recently, dual-branch codecs \cite{lu2024hybridflow, xue2025dlf} have been introduced to support more compact and flexible latent representations.
While these methods highlight the strong feature learning capacity of VAE-based codecs, their reconstruction quality at extremely low bitrates remains limited by model capacity and training scale.

\textbf{Diffusion-based Generative Image Compression.} 
Diffusion models have rapidly evolved in recent years, progressing from theoretical foundations~\cite{sohl2015deep, ho2020denoising} to practical high-quality implementations~\cite{rombach2021highresolution}.
Building on these advances, recent studies \cite{lei2023text+,careil2023towards,li2024towards,vonderfecht2025lossy, xu2025} have explored the use of generative priors from large-scale pretrained diffusion models in image compression, demonstrating improved perceptual realism compared to conventional VAE-based approaches.
Eric et al. \cite{lei2023text+} combine pretrained diffusion with ControlNet \cite{zhang2023adding} to reconstruct images from captions and compressed sketches. 
PerCo \cite{careil2023towards} fine-tunes diffusion models using captions and quantized features, with captions generated by a large BLIP2 model \cite{li2023blip}. 
DiffEIC \cite{li2024towards} conditions diffusion models on VAE latents, showing strong performance even without text, and their subsequent work \cite{li2024diffusion} improves encoder and introduces relay fine-tuning. 
DiffC \cite{vonderfecht2025lossy} demonstrates zero-shot compression with diffusion models using reverse-channel coding. 
DDCM \cite{ohayon2025compressed} replaces the continuous Gaussian noise in sampling with discrete codebook selections for compression task.
While these approaches improve realism at ultra-low bitrates, they often struggle to maintain high fidelity. 
Moreover, multi-step sampling in DDIM \cite{song2020denoising} introduces substantial decoding overhead, limiting their practicality. 
By contrast, our OneDC achieves high fidelity and realism with significantly faster decoding via one-step sampling. 
While prior work \cite{li2024towards} suggests high-level semantics (e.g., text) may be optional, we find such guidance remains crucial in our one-step diffusion for image decoding task, as detailed in Section~\ref{sec:4.3.1}.

\textbf{One-step Diffusion Models.} 
To reduce the computational cost, recent efforts have focused on distilling multi-step models into efficient one-step generators. 
Song et al.~\cite{song2023consistency} introduce Consistency Models, which achieve few-step generation by enforcing self-consistency along diffusion trajectories.
Yin et al. \cite{yin2024one} propose Distribution Matching Distillation (DMD), which minimizes an approximate KL divergence between real and generated image distributions modeled by multi-step teacher networks. 
Their subsequent work, DMD2 \cite{yin2024improved}, further improves generation quality by introducing a diffusion-GAN framework and enhanced training strategies. 
Based on DMD, Song et al. \cite{song2024multi} introduce class-specific student models to improve performance across diverse categories. 
SwiftBrush series \cite{nguyen2024swiftbrush, dao2024swiftbrush2} adopt a LoRA adaptation \cite{hu2022lora} for variational score distillation and further optimize diversity by integrating a clamped CLIP loss.
Independently, Kang et al. \cite{kang2024distilling} employ perceptual loss supervision and a multi-scale U-Net discriminator for one-step distillation. 
These models have also been extended to low-level vision tasks such as super-resolution \cite{dong2024tsd, wu2024one, wang2024sinsr, kim2024tddsr} and image restoration \cite{guo2025compression}, highlighting the potential of one-step diffusion beyond generation. 
Inspired by these advances, we explore its application in generative image compression, a setting that remains underexplored.

%% file: section_3.tex
\section{Methodology}

\begin{figure}[t]
    \centering
    \includegraphics[width=1.0\linewidth]{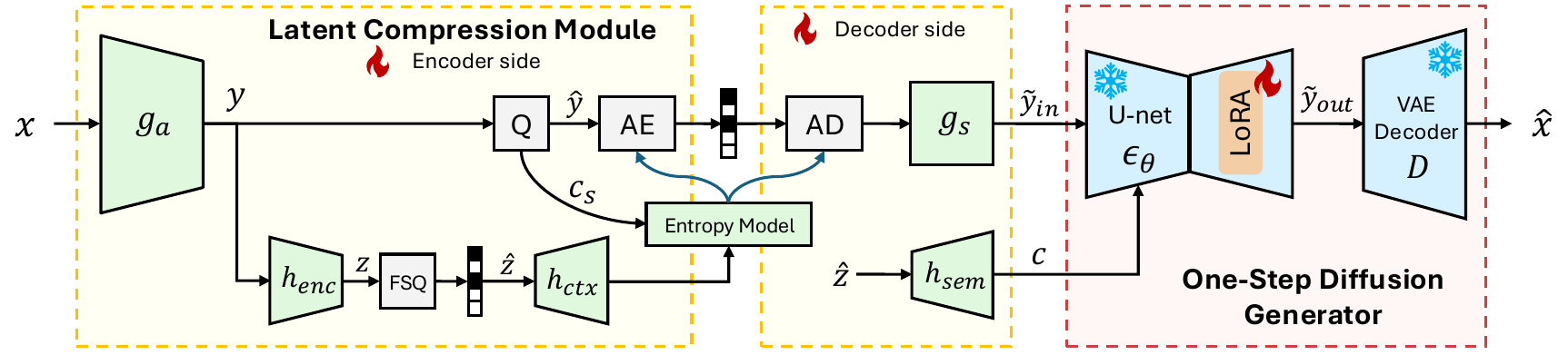}
    \vspace{-5mm}
    \caption{Overview of the OneDC framework. Q denotes scalar quantization, and FSQ stands for finite scalar quantization. AE and AD refer to the arithmetic encoder and decoder, respectively. $h_{ctx}$ and $h_{sem}$ represent the context and semantic decoders used in the hyperprior branch.}
    \label{fig:3}
    \vspace{-5mm}
\end{figure}

We propose \textbf{OneDC}, a one-step diffusion-based generative image compression framework that achieves high-quality reconstruction with low-latency decoding under ultra-low bitrate constraints.
Section~\ref{sec:3.1} introduces the coding pipeline of our method, consisting of a latent compression module that encodes the input image into a compact latent representation and a one-step diffusion generator that synthesizes reconstructions with realistic details.
Given the critical role of semantic guidance in our one-step diffusion and the limitations of textual prompts in compression scenarios, Section~\ref{sec:3.2} introduces the use of the hyperprior as an alternative conditioning signal.
To further tap the potential of the hyperprior, we propose a semantic distillation strategy that transfers knowledge from a pretrained generative tokenizer to the hyperprior codec, thereby improving semantic accuracy.
Finally, Section~\ref{sec:3.3} presents our training scheme, which combines pixel-domain and latent-domain objectives to jointly optimize reconstruction fidelity and perceptual realism.

\subsection{Framework Overview}
\label{sec:3.1}

Fig.~\ref{fig:3} shows the overview of our OneDC framework. On the encoder side, an analysis transform encodes the input $x$ into a compact latent representation $y=g_a(x)$, which is then quantized to $\hat{y}$. A hyper encoder further processes the latent into a hyperprior $z=h_{enc}(y)$, quantized as $\hat{z}$. Then, the entropy model takes decoded hyperprior context $c_h = h_{ctx}(\hat{z})$ and spatial context $c_s$ to predict the distribution of $\hat{y}$, enabling bitrate estimation during training and entropy coding during inference.

\begin{figure}[t]
    \centering
    \includegraphics[width=1.0\linewidth]{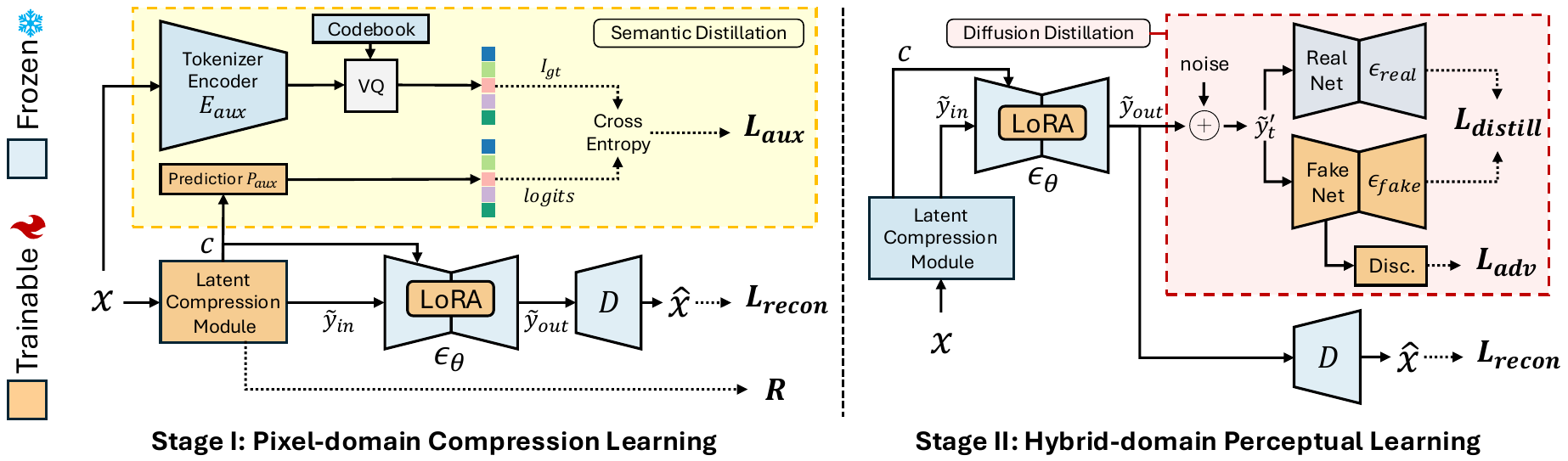}
    \vspace{-3mm}
    \caption{Two stage training pipeline of OneDC. The codebook in semantic distillation is initialized from the pretrained tokenizer, and the discriminator in diffusion distillation is abbreviated as Disc.}
    \label{fig:4}
    \vspace{-2mm}
\end{figure}

On the decoder side, the received $\hat{y}$ is passed through a synthesis transform to produce an initial latent $\tilde{y}_{in}=g_s(\hat{y})$ for generation. 
Given the distinct roles of entropy modeling and semantic processing, where the former estimates low-dimensional distribution parameters and the latter represents high-dimensional visual contents, we introduce an additional semantic decoder for the hyperprior to better address their different requirements.
It extracts semantic guidance $c = h_{sem}(\hat{z})$ from the quantized hyperprior, which is injected into the cross-attention layers of the one-step diffusion generator.
The diffusion generator refines the initial latent in only one step: $\tilde{y}_{out} = \epsilon_{\theta}(\tilde{y}_{in}, c)$, which is then decoded by a pretrained VAE decoder to produce the final reconstruction $\hat{x} = D(\tilde{y}_{out})$.

\subsection{Semantic Guidance with Hyperprior}
\label{sec:3.2}
\subsubsection{From Text to Hyperprior}

In the one-step diffusion setting, semantic guidance (the input of the cross-attention layers \cite{rombach2021highresolution}) plays an even more critical role than in conventional multi-step diffusion. Unlike multi-step methods, which iteratively refine the output and can gradually correct semantic inconsistencies, one-step diffusion models rely entirely on a single forward pass, making accurate semantic conditioning essential. 
While existing diffusion models typically rely on textual prompts to provide high-level semantic guidance, such prompts are suboptimal in the context of natural image compression.
Text struggles to capture localized semantics and often fails to describe fine-grained visual attributes—such as object boundaries and textures—particularly in high-resolution imagery.
In addition, generating text descriptions typically requires large models \cite{li2023blip}, introducing significant computational overhead.

\begin{figure*}[t]
    \centering
    \includegraphics[width=1.0\linewidth]{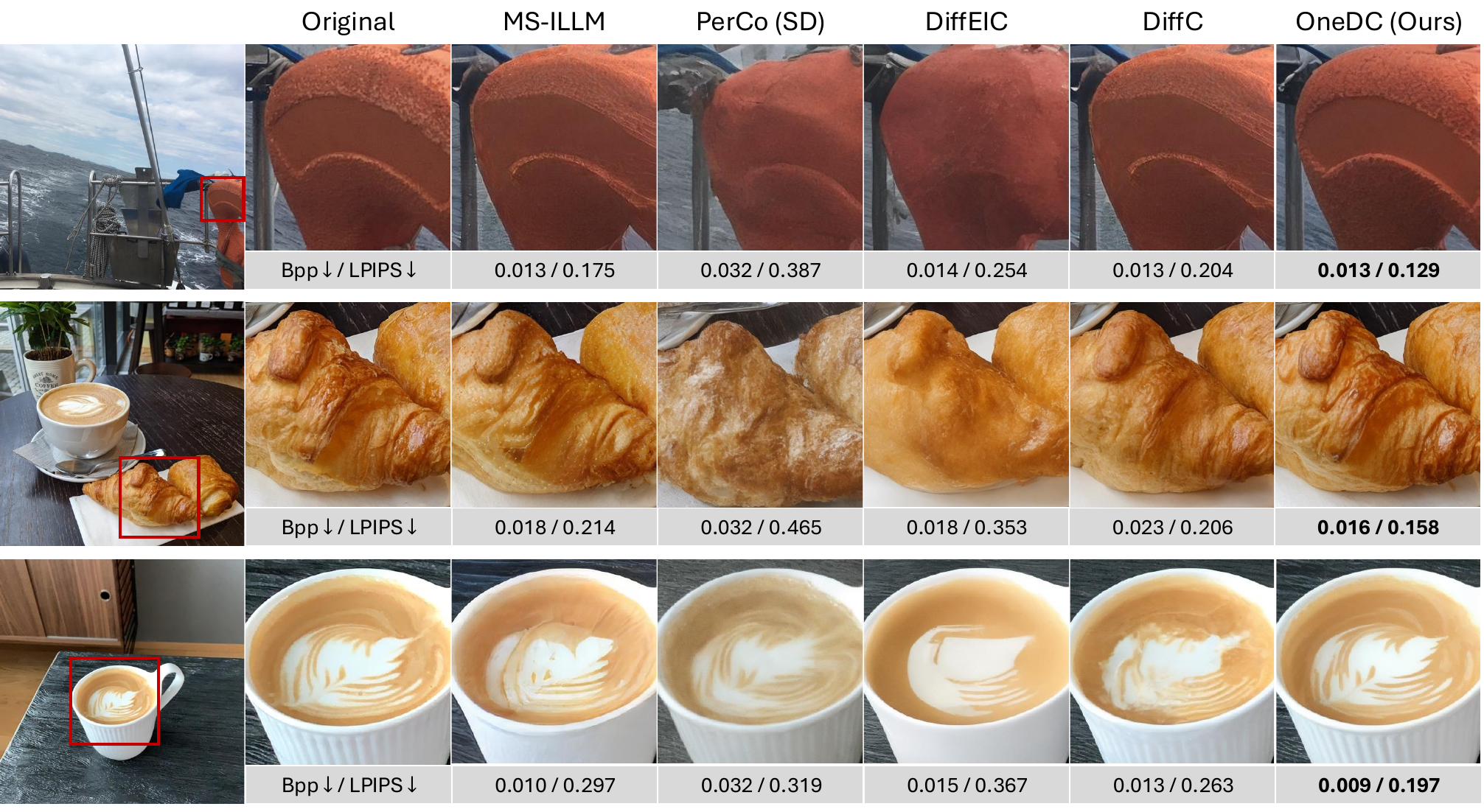}
    \vspace{-5mm}
    \caption{Visual examples on the CLIC2020 test set. Zoom in for better view.}
    \vspace{-4mm}
    \label{fig:visual1}
\end{figure*}

Recent work~\cite{jia2024generative} suggests that the categorical distributed hyperprior with vector quantization (VQ) can capture semantic information.
Building on this insight, OneDC replaces textual prompts with the categorical hyperprior $\hat{z}$ as semantic guidance for the diffusion model.
To compress the hyperprior, we adopt finite scalar quantization (FSQ)~\cite{mentzer2023finite}, a simpler yet more effective alternative to VQ~\cite{esser2021taming}, to learn a categorical $\hat{z}$ with semantic information.
The hyperprior $z$ is compressed into 7 channels with 4 quantization levels each, yielding an equivalent codebook size of 16,384. 
At a $64\times$ spatial downsampling, this design results in a bitrate of just 0.0034 bpp, incurring very low transmission overhead.
To substitute for text embeddings, we introduce a semantic decoder $h_{sem}$ that transforms the quantized hyperprior $\hat{z}$ into a semantic context $c \in \mathbb{R}^{B \times N \times D}$, where $N=H' \times W'$ is the flattened spatial shape and $D$ is the embedding dimension. This context is injected into the cross-attention layers of the one-step diffusion U-Net, replacing the textual embeddings used in the original design.
In each cross-attention layer, the input latent feature $f_{in}$ serves as the query, while the semantic context $c$ provides both keys and values:

\vspace{-3mm}
\begin{equation}
\label{eq:0}
f_{out} = \mathrm{Softmax}\left(\frac{QK^\top}{\sqrt{d_{k}}}\right)V, \quad\text{where}~ Q=W_{Q}f_{in}, ~ K=
W_{k}c, ~ V=W_{v}c
\end{equation}
\vspace{-3mm}

where $d_k$ denotes the dimensionality of the key vectors and $f_{out}$ is the output feature of the cross-attention layer. This design enables every spatial location in the latent to attend adaptively to semantically relevant tokens, strengthening content alignment. Compared with text conditioning, hyperprior-based guidance offers three advantages:
(1) As shown in Fig.~\ref{fig:2}, conditioning the one-step diffusion model on categorical hyperprior tokens yields more faithful reconstructions than using text prompts, validating their effectiveness as semantic guidance.
(2) The $64\times$ downsampled hyperprior provides a large receptive field while retaining spatial locality, thereby offering a more balanced global–local semantics compared to purely global text embeddings.
(3) Unlike separately trained text encoders \cite{radford2021learning}, our approach enables end-to-end optimization of semantic guidance, allowing the hyperprior to adapt jointly with the diffusion model for improved content expression.

\subsubsection{Semantic Distillation for Hyperprior}

While the hyperprior effectively captures semantic cues for diffusion guidance, its representation capability can be further improved.
In particular, we propose a semantic distillation mechanism aimed at fully tapping the potential of the hyperprior branch.
This distillation guides the hyperprior encoder to better capture visual content and improve the decoding quality of the semantic decoder $h_{sem}$.
Specifically, we transfer knowledge from a pretrained high-capacity generative tokenizer~\cite{esser2021taming} to the hyperprior codec via an auxiliary distillation task.
This choice is motivated by the structural similarity between the categorical hyperprior codec and the generative tokenizers that are known to produce semantically rich discrete representations~\cite{yu2023spae}.
By leveraging this compatibility, we use the tokenizer as a teacher to guide the hyperprior toward more effective semantic encoding.

Following GLC~\cite{jia2024generative}, we introduce a transformer-based predictor $P_{aux}$~\cite{zhou2022towards}, trained to predict discrete token labels from the hyperprior-derived semantics $c$. 
The ground truth labels $I_{gt}$ are obtained from a pretrained tokenizer encoder $E_{aux}$. 
Since the tokenizer is optimized for perceptual reconstruction, its objectives align well with OneDC, making it an effective teacher.
Moreover, the smaller information bottleneck in the hyperprior branch naturally filters out redundant information, distilling the most salient semantic features from the teacher.
The distillation is supervised via cross-entropy (CE) loss:

\vspace{-3mm}
\begin{equation}
\label{eq:1}
    I_{gt} = VQ(E_{aux}(x)), \;\;\;\; L_{aux}=CE(I_{gt}, P_{aux}(c))
\end{equation}
\vspace{-4mm}

Both $P_{aux}$ and $E_{aux}$ are used only during training, introducing no inference overhead. 
As shown in Fig.\ref{fig:2}, our semantic distillation enhances content fidelity, and ablation results in Section\ref{sec:4.3.1} further validate its effectiveness in improving final reconstruction quality.

\begin{figure*}[t]
    \centering
    \includegraphics[width=1.0\linewidth]{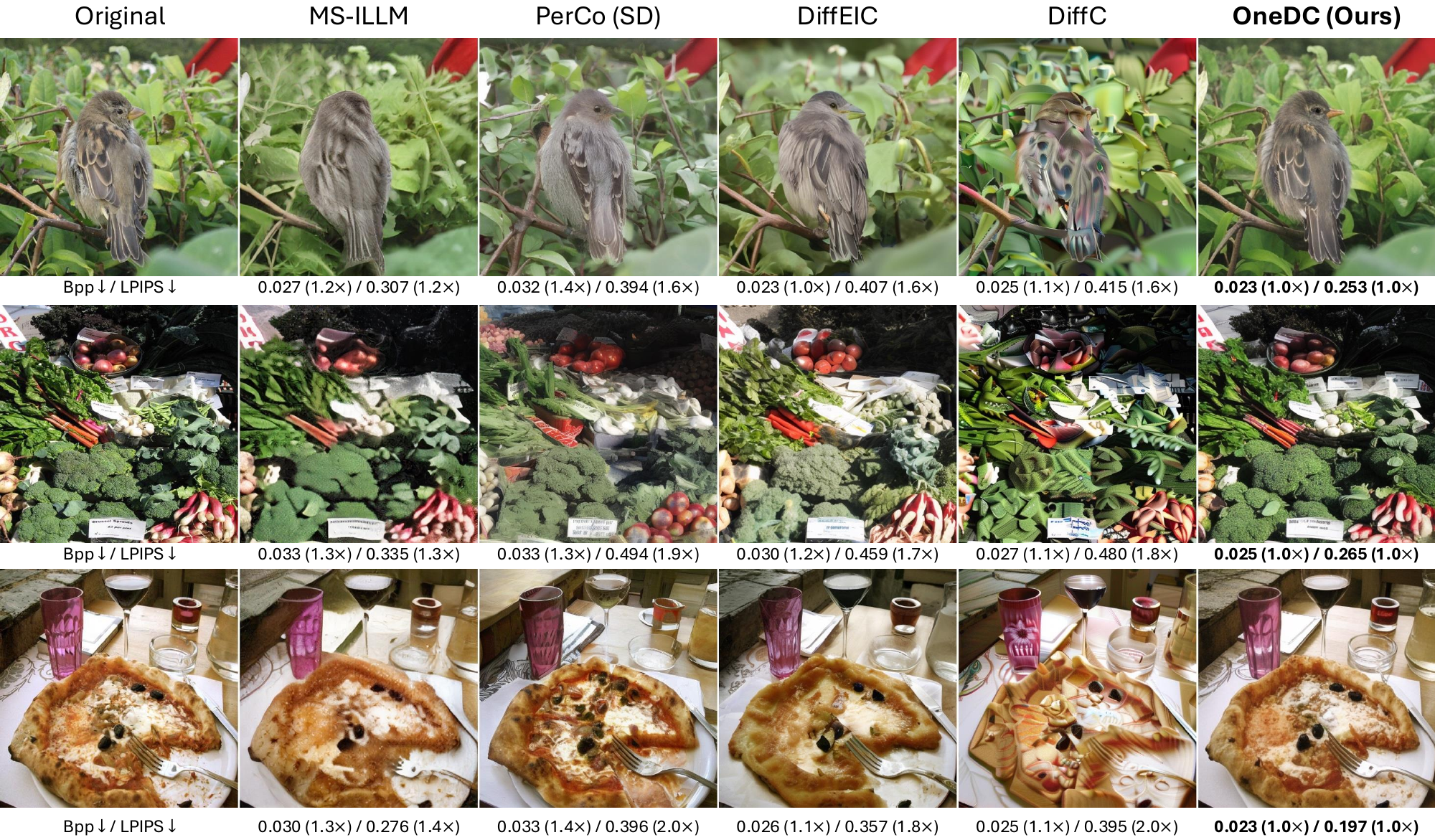}
    \vspace{-3mm}
    \caption{Visual examples on the MS-COCO 30K dataset. Zoom in for better view.}
    \vspace{-2mm}
    \label{fig:visual2}
\end{figure*}

\subsection{Training Strategy}
\label{sec:3.3}
To enhance training efficiency and stability, we adopt a two-stage training strategy, inspired by prior works \cite{mentzer2020high, muckley2023improving}.
As shown in Fig.~\ref{fig:4}, stage I mainly trains the compression module while stage II focuses on fine-tuning the one-step diffusion model for better reconstruction quality.
In both stages, the one-step diffusion model is adapted via LoRA layers \cite{hu2022lora}, allowing fast convergence for the compression task while preserving rich generative priors.

\textbf{Stage I: Pixel-domain Compression Learning}
This stage aims to: (1) train the codec to compress images into compact latent representations with high-fidelity reconstruction, (2) embed semantic information into the hyperprior via distillation, and (3) adapt the one-step diffusion model to synthesize fine-grained details on the decoder side. The total loss is defined as: 

\vspace{-3mm}
\begin{equation}
\label{eq:2}
    L_{stageI} = L_{recon} + \lambda R + \alpha L_{aux}, \;\;\;\; \text{where} \;\; L_{recon}=L_{1}(x,\hat{x})+L_{perceptual}(x,\hat{x})
\end{equation}
\vspace{-4mm}

We use LPIPS \cite{zhang2018unreasonable} as the perceptual loss and $L_{1}$ as the pixel loss. $R$ is the bitrate loss from the quad-tree-based spatial entropy model \cite{Li_2023_CVPR,li2024neural}, and $L_{aux}$ is the semantic distillation loss (Eq.~\ref{eq:1}). The hyperparameter $\lambda$ controls the compression ratio and $\alpha$ is the weight of semantic distillation loss.

\begin{figure*}[t]
    \centering
    \includegraphics[width=1.0\linewidth]{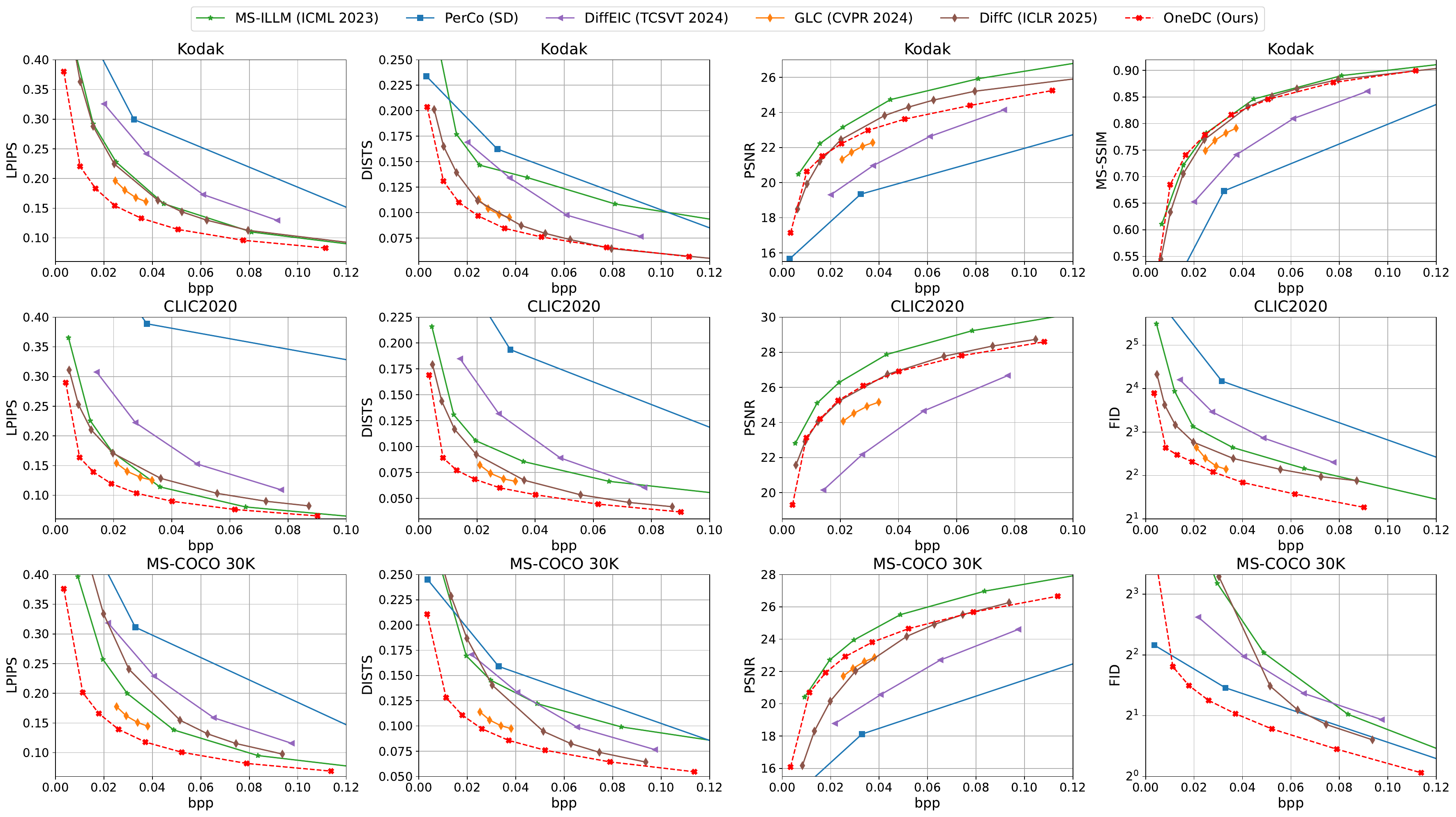}
    \vspace{-5mm}
    \caption{Rate–distortion curves under the \textit{full-resolution} setting. Zoom in for better view.}
    \vspace{-2mm}
    \label{fig:exp1}
\end{figure*}

\begin{figure*}[t]
    \centering
    \includegraphics[width=1.0\linewidth]{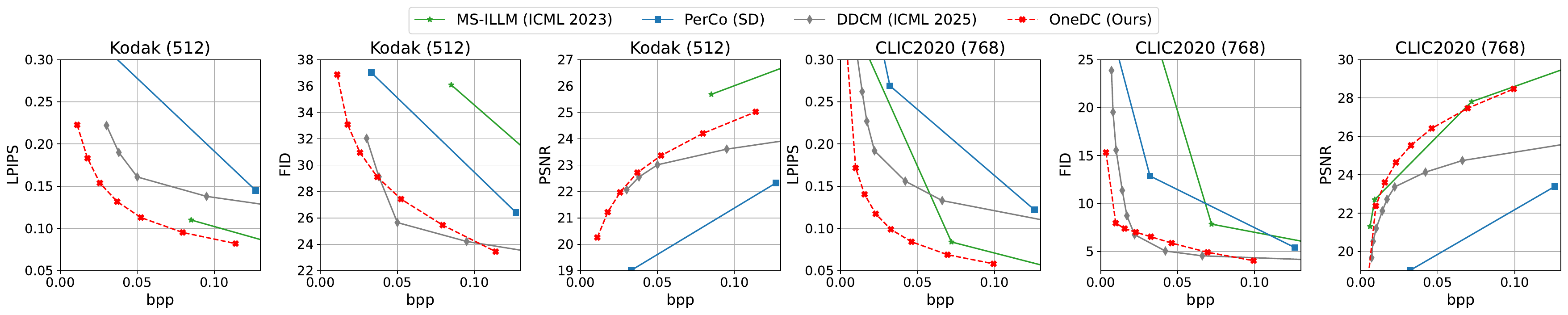}
    \vspace{-5mm}
    \caption{Rate–distortion curves under the \textit{resize \& center-crop} on Kodak 512$\times$512 and CLIC2020 768$\times$768. Zoom in for better view.}
    \vspace{-5mm}
    \label{fig:exp2}
\end{figure*}

\textbf{Stage II: Hybrid-domain Perceptual Learning.} 
This stage fine-tunes the one-step diffusion model to enhance perceptual realism without degrading fidelity. 
The latent compression module is fixed to improve training stability \cite{mentzer2020high}. 
Prior work has shown that pixel-level optimization alone is insufficient for optimal perceptual quality \cite{jia2024generative}. Thus, we adopt a joint training strategy in both pixel and latent domains.
Specifically, we integrate the diffusion-based distillation method from \cite{yin2024improved}, which transfers generation knowledge from a pretrained multi-step diffusion teacher. The training also involves both perceptual supervision from the pixel domain and adversarial alignment in the latent domain, which ensures fidelity by aligning reconstruction with the original image. The total loss is:

\vspace{-7mm}
\begin{gather}
    L_{stageII} = L_{distill} + \beta L_{recon} + \gamma L_{adv}, \;\;\;\; \text{where:} \\
    L_{distill} = \mathbb{E}_{t,\tilde{y}'_{t}}[\epsilon_{fake}(\tilde{y}'_{t},t)-\epsilon_{real}(\tilde{y}'_{t},t)], \;\;  L_{adv} = \mathbb{E}_{t,\tilde{y}'_{t}}[-Disc(\epsilon_{fake}(\tilde{y}'_{t},t), t)]
\end{gather}
\vspace{-3mm}

Here, $\epsilon_{real}$ and $\epsilon_{fake}$ denote the real and fake score networks in the teacher model, and $\tilde{y}'_t$ is the noised latent at timestep $t$. 
$Disc$ refers to the discriminator.
$\beta$ and $\gamma$ balance the pixel and adversarial losses. 
Additional implementation and loss details are provided in the supplementary material.

%% file: section_4.tex
\section{Experiment}

\subsection{Implementation and Evaluation Settings}

\textbf{Model and Training.}
Our generator adopts the U-Net architecture from the Stable Diffusion 1.5 (SD1.5) \cite{rombach2021highresolution} and is initialized with weights from a pretrained one-step text-to-image task \cite{yin2024improved}.
In Stage I, we use the pretrained tokenizer from MaskGIT \cite{chang2022maskgit} for semantic distillation.
In Stage II, we employ the multi-step SD1.5 pretrained model as the teacher for diffusion distillation.
To improve high-resolution adaptability, we randomly crop patches of size 512 or 1024 during training. Models are optimized using AdamW \cite{loshchilov2017decoupled}. Additional settings are provided in the supplementary material.

\textbf{Compared methods.}
We compare our OneDC with generative image codecs at a low-bitrate scenario. The compared methods include the vanilla VAE-based generative codec MS-ILLM \cite{muckley2023improving} and GLC \cite{jia2024generative}, and the recent diffusion-based methods: PerCo (SD) \cite{korber2024perco}, DiffEIC \cite{li2024towards}, DiffC \cite{vonderfecht2025lossy} (SD2.1-based) and DDCM \cite{ohayon2025compressed}. 
For most methods, we evaluate at \textit{full-resolution}, consistent with previous implementations in generative neural codecs \cite{mentzer2020high, muckley2023improving, jia2024generative}. For some diffusion-based methods like DDCM, we additionally provide results under the \textit{resize \& center-crop} setting, following DDCM protocol \cite{ohayon2025compressed} for fair comparison. Further details are provided in the supplementary material.

\textbf{Evaluation datasets and metrics.} 
We evaluate OneDC on several datasets, including Kodak \cite{kodak}, CLIC2020 test set \cite{CLIC2020}, and MS-COCO 30K \cite{lin2014microsoft}. Reconstruction fidelity is evaluated using perceptual metrics LPIPS~\cite{johnson2016perceptual} and DISTS~\cite{ding2020image}, along with the traditional metrics PSNR and MS-SSIM \cite{wang2003multiscale}, while generative realism is measured by the no-reference perceptual metric FID~\cite{heusel2017gans}. Bitrate saving is measured by BD-Rate \cite{bjontegaard2001calculation}. It is worth noting that FID is computed on overlapping $256 \times 256$ patches at \textit{full-resolution} CLIC2020, following previous practice~\cite{muckley2023improving, jia2024generative}. For the MS-COCO 30K dataset with $512 \times 512$ images, we evaluate FID on full-images, consistent with \cite{careil2023towards, korber2024perco}. However, large patch leads to unreliable FID on the small test set (Kodak with 24 images) \cite{careil2023towards}. To address this, under the \textit{resize \& center-crop} setting we follow the DDCM ~\cite{ohayon2025compressed} and compute FID on overlapping $64 \times 64$ patches for Kodak. More details are provided in the supplementary material.

\subsection{Main Results}

\textbf{Quantitative Evaluation.}
Fig.~\ref{fig:exp1} presents the \textit{full-resolution} comparison between OneDC and existing methods across multiple distortion metrics.
Across all datasets, including high-resolution CLIC2020, low-resolution Kodak and MS-COCO 30K, OneDC consistently achieves the best reference-based perceptual fidelity, as measured by LPIPS and DISTS.
Compared with the previous SOTA, multi-step diffusion-based DiffC, OneDC achieves \textbf{55.27\%} bitrate saving on Kodak and \textbf{54.60\%} on CLIC2020 in terms of LPIPS, demonstrating superior compression efficiency.
On traditional metrics such as PSNR, OneDC also exhibits competitive performance.
For generation realism, OneDC attains lower FID than all multi-step diffusion baselines when FID is evaluated on the CLIC2020 test set.
On the MS-COCO 30K dataset, OneDC provides a \textbf{39.55\%} bitrate saving over PerCo (SD), previously the highest-realism method on this dataset. Such result further underscoring our method's effectiveness.
Furthermore, Fig.~\ref{fig:exp2} reports results under the \textit{resize \& center-crop} setting: OneDC achieves better LPIPS and PSNR than DDCM while maintaining competitive realism, demonstrating strong fidelity-oriented perceptual compression.
In conclusion, OneDC delivers faithful and perceptually compelling reconstructions across a wide range of resolutions and content complexities, despite relying on a single sampling step.
These evaluation results support our hypothesis that iterative sampling is unnecessary for high-quality diffusion-based generative compression.

\textbf{Qualitative Evaluation.}
Fig.~\ref{fig:visual1} presents visual comparisons on the CLIC2020 test set (\textit{full-resolution}).
At the lowest bitrate, OneDC produces the most natural and faithful reconstructions with sharp high-frequency details, while MS-ILLM yields noticeably blurred textures and multi-step diffusion-based methods (DiffEIC, PerCo, and DiffC) introduce structural distortions and inconsistent details.
Similar trends are observed on the MS-COCO 30K dataset (Fig.~\ref{fig:visual2}), where MS-ILLM remains blurry and multi-step diffusion methods struggle to preserve visual fidelity.
These qualitative results, together with the quantitative comparisons, highlight the SOTA performance of the proposed OneDC.
Additional visual examples are provided in the supplementary material.

\begin{table}[t]
\begin{minipage}[t]{0.43\textwidth}
\captionof{table}{\small Ablation studies with BD-Rate ($\%$) $\downarrow$.}
\vspace{1mm}
\centering
\renewcommand\arraystretch{1.1}
\setlength{\tabcolsep}{3.5pt}
\label{tab:1}
\resizebox{0.955\textwidth}{!}{
\begin{tabular}{lcc}
\toprule
\multicolumn{1}{l}{\multirow{2}{*}{Settings}} & \multicolumn{2}{c}{CLIC2020} \\ [2pt]
  & DISTS & FID  \\ \midrule
\textcolor{gray}{\emph{Semantic guidance}}  \\
No guidance                                     & 44.0 & 45.1 \\
Text guidance                                   & 24.2 & 36.3 \\
Hyperprior guidance                             & 20.7 & 24.3 \\
Hyperprior + Sem. Distil.$\to$ \textbf{Ours}    & 0.00 & 0.00 \\ \midrule
\textcolor{gray}{\emph{Loss variation}}  \\
Pixel-domain only                     & 11.4  & 51.8 \\
Latent-domain only                    & 60.7  & 37.1 \\
Hybrid-domain $\to$ \textbf{Ours}     & 0.00  & 0.00 \\ \bottomrule
\end{tabular}
}
\end{minipage}
\hfill
\begin{minipage}[t]{0.53\textwidth}
\captionof{table}{\small Comparison of coding time and BD-Rate ($\%$) $\downarrow$.}
\vspace{1mm}
\centering
\renewcommand\arraystretch{1.2}
\setlength{\tabcolsep}{3.5pt}
\label{tab:2}
\resizebox{1.0\textwidth}{!}{
\begin{tabular}{lccccc}
\toprule
\multirow{2}{*}{Methods} & \multicolumn{2}{c}{Times (s)} & \multicolumn{3}{c}{MS-COCO 30K} \\ [2pt]
                         & Enc.  & Dec.  & LPIPS  & DISTS  & FID     \\ \midrule
\textcolor{gray}{\emph{VAE-based}} & & & & &        \\
MS-ILLM                  & 0.14   & 0.17     & 138.3   & 253.0  & 478.4    \\ \midrule
\textcolor{gray}{\emph{Multi-step diffusion}} & & & & &   \\
DiffEIC                  & 0.32   & 12.4    & 305.0  & 239.1  & 341.0  \\
PerCo (SD)               & 0.58   & 8.80    & 538.8  & 345.8  & 59.6   \\ 
DiffC                    & 3.9$\sim$15.6    & 6.9$\sim$10.8 & 234.0  & 196.1  & 690.9  \\ \midrule
\textcolor{gray}{\emph{One-step diffusion}} & & & & &       \\
OneDC $\to$ \textbf{Ours} & 0.15   & 0.34   & 0.00    & 0.00   & 0.00    \\ \bottomrule
\end{tabular}
}
\end{minipage}

\vspace{2mm}
\noindent\parbox{\textwidth}{\footnotesize
Notes: Metrics in the tables denote those used for BD-Rate calculation \cite{bjontegaard2001calculation}; \textbf{Ours} is the anchor (0.00\%).
}

\vspace{-1mm}
\end{table}

\begin{figure}[t]
    \centering
    \includegraphics[width=1.0\linewidth]{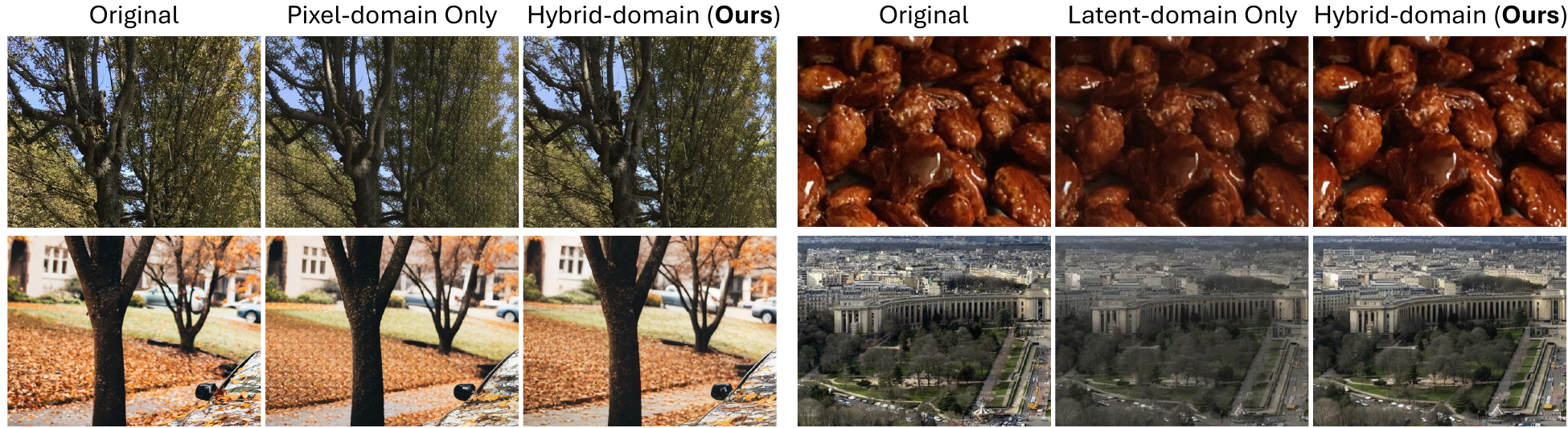}
    \caption{Visual examples of the ablation study in hybrid-domain perceptual learning. No latent-domain training leads artifacts (Left), while no pixel-domain supervision leads color shift (Right).}
    \label{fig:exp3}
    \vspace{-4mm}
\end{figure}

\subsection{Ablation Studies and Efficiency Evaluation}
We conduct ablation studies to investigate the effectiveness of semantic guidance and hybrid-domain perceptual learning. Models are evaluated on the CLIC2020 test set with BD-Rate \cite{bjontegaard2001calculation}.

\textbf{Semantic guidance.}
\label{sec:4.3.1}
We first validate the importance of semantic guidance and then evaluate the effectiveness of our proposed semantic distillation for the hyperprior codec.
As shown in Table~\ref{tab:1}, removing semantic guidance (“No Guidance”) significantly degrades reconstruction quality—by up to 40\%—highlighting its critical role in our one-step diffusion model.
While text-based guidance improves performance, it underperforms the hyperprior guidance on high-resolution CLIC2020 images. 
This suggests that the hyperprior's locally captured semantics offer a more precise and spatially aligned representation of the image content.
Further applying semantic distillation to the hyperprior (“Hyperprior + Sem. Distill.”) leads to additional gains, indicating that transferring prior knowledge from a pretrained tokenizer enhances the semantic capability of our hyperprior branch.

\textbf{Hybrid-domain Perceptual Learning.} 
We evaluate the effectiveness of hybrid pixel-latent domain training in Stage II, as summarized in Table~\ref{tab:1}.
(1) Using the pixel-domain loss only (\textit{No} $L_{distill}$ and $ L_{adv}$) leads to notable drops in both FID and DISTS scores. 
Visual examples in Fig.~\ref{fig:exp3}-Left show grid-like artifacts, reflecting a significant decline in perceptual quality. 
This underscores the importance of diffusion distillation in enforcing distributional alignment for realistic reconstruction.
(2) Using the latent-domain loss only (\textit{No} $L_{recon}$) also results in consistent degradation across all metrics, accompanied by visible color shifts in the reconstructions (Fig.~\ref{fig:exp3}-Right). 
This highlights the importance of pixel-level supervision in image compression, which emphasizes accurate reconstruction—unlike the pure image generation task that focuses solely on visual realism.

\textbf{Operation Efficiency.}
We evaluate the coding times of different methods on $1024\times1024$ images using an A100 GPU.
These runtime results are presented alongside BD-Rate results on the MS-COCO 30K dataset to offer a comprehensive view of operational efficiency.
As shown in Table~\ref{tab:2}, OneDC achieves a significant decoding speed advantage—over $20\times$ faster—compared to multi-step diffusion-based methods, while still maintaining impressive reconstruction quality.
Although OneDC is slightly slower than the pure VAE-based MS-ILLM due to its larger model, this added capacity is crucial for preserving perceptual quality under extreme compression, as evidenced by its superior DISTS and FID performance.
Overall, these results underscore the effectiveness of OneDC in balancing fidelity, realism, and efficiency at ultra-low bitrates.

\section{Conclusion}
We present OneDC, a novel one-step diffusion-based generative image codec designed for high-quality and efficient compression under ultra-low bitrate scenarios.
OneDC integrates a latent compression module for efficient coding and a one-step diffusion generator for fast reconstruction.
To provide the critical semantic guidance for the one-step diffusion model and address the limitations of textual guidance, we leverage the hyperprior as a substitute.
Furthermore, we improve the semantic capability of the hyperprior through a distillation mechanism.
To enhance training efficiency and stability, we adopt a two-stage strategy: Stage I focuses on compression training through pixel-domain supervision, while Stage II refines reconstruction quality via hybrid-domain perceptual learning.
Extensive experiments demonstrate that OneDC achieves SOTA perceptual quality with fast decoding, highlighting the strong potential of one-step diffusion models in generative image compression.

\textbf{Limitation.} Although OneDC offers substantial speedups over multi-step diffusion methods, its decoding speed does not meet real-time requirements. In future work, we plan to explore model distillation and architectural optimization to enhance efficiency.

%% file: section_6_append.tex
\section{Experiment}
\begin{figure}[t]
    \centering
    \includegraphics[width=\linewidth]{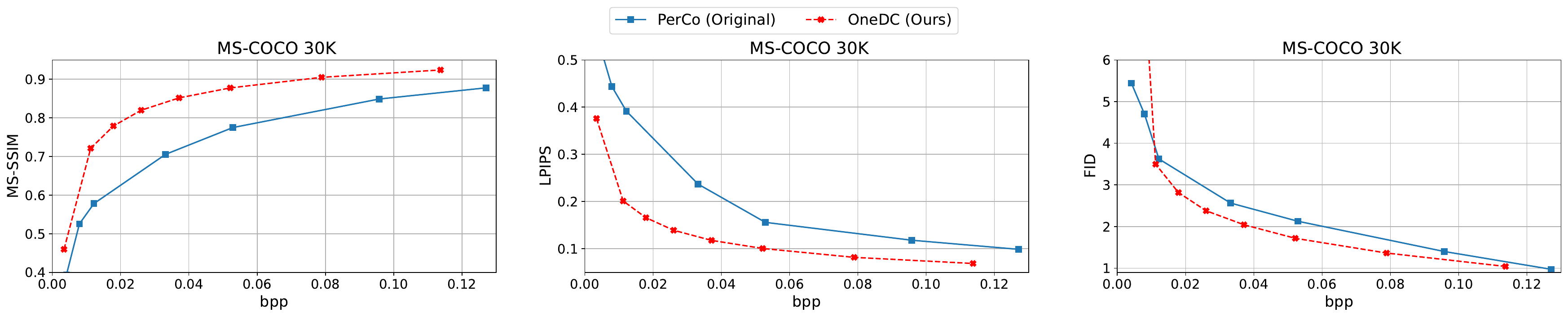}
    \vspace{-5mm}
    \caption{Comparison with original PerCo \cite{careil2023towards} on the MS-COCO 30K dataset.}
    \label{fig:supp_exp0}
    \vspace{-1mm}
\end{figure}

\textbf{Evaluation of third-party models.} 
We evaluate MS-ILLM~\cite{muckley2023improving} using the official checkpoints and fine-tune them with the provided code to support lower bitrates.
For GLC~\cite{jia2024generative}, we report the results directly from its paper because the code and models are not publicly available at present.
For DiffEIC~\cite{li2024towards} and DiffC~\cite{vonderfecht2025lossy}, we use the official implementations and released models. 
Specifically, we adopt the Stable Diffusion (SD) 2.1-based variant of DiffC, which shows slightly better performance compared to the SD 1.5 version.
For PerCo~\cite{careil2023towards}, we use a publicly available reimplementation PerCo (SD)~\cite{korber2024perco}, as the original code has not been released. We also report the comparison with original PerCo by extract data from their paper (Fig. \ref{fig:supp_exp0}).
For DDCM \cite{ohayon2025compressed}, we use the number reported in their paper.
For multi-step diffusion codecs, we follow the default sampling settings provided in their code (e.g., 50 for DiffEIC, 20 for PerCo).

\textbf{Test settings and FID calculation.} 
At the \textit{full-resolution} setting, we compute FID using overlapping $256 \times 256$ patches for the CLIC2020 and DIV2K test sets~\cite{div2k}, following the protocol of~\cite{mentzer2020high}. For the MS-COCO 30K dataset with $512 \times 512$ images, FID is evaluated on entire images, consistent with~\cite{careil2023towards, korber2024perco}.
At the \textit{resize \& center-crop} setting, we resize the short side of each image (512 for Kodak, 768 for CLIC2020 test set) and then apply a center crop. In this setting, we use $64 \times 64$ patches for FID calculation on Kodak and $128 \times 128$ patches on CLIC2020 test set, consistent with DDCM~\cite{ohayon2025compressed}.
It is worth noting that DiffEIC also evaluates under the \textit{resize \& center-crop} setting on CLIC2020 test set at 768 resolution, but computes FID with overlapping $256 \times 256$ patches. For completeness, we additionally report results under DiffEIC’s protocol, denoted as \textit{resize \& center-crop, 256 FID}. 

\begin{figure}[t]
    \centering
    \includegraphics[width=\linewidth]{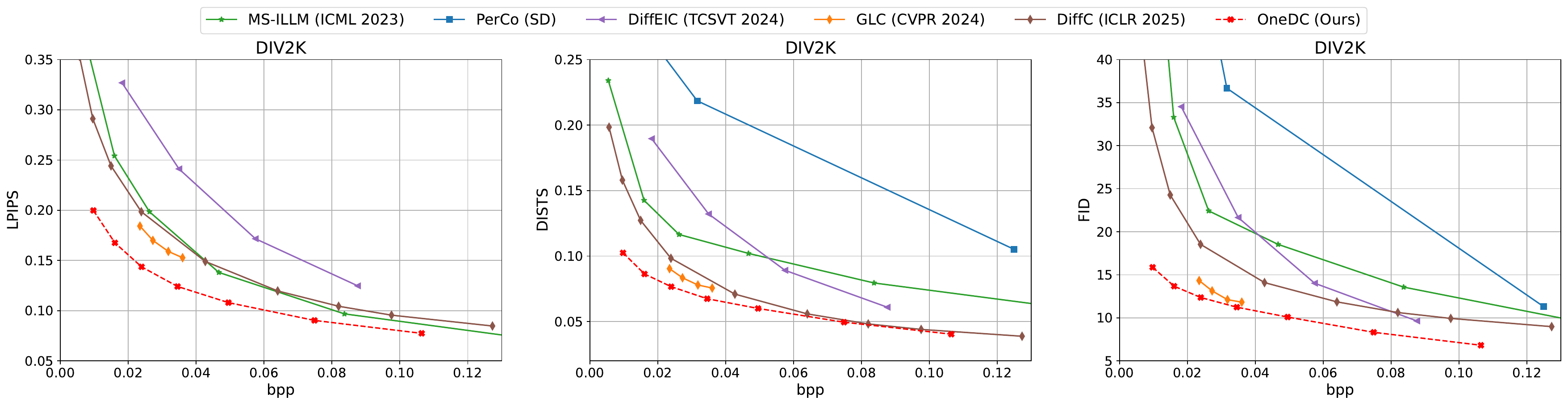}
    \vspace{-5mm}
    \caption{Rate–distortion curves on the DIV2K dataset at \textit{full-resolution} setting.}
    \label{fig:supp_exp1}
    \vspace{-1mm}
\end{figure}

\begin{figure}[t]
    \centering
    \includegraphics[width=\linewidth]{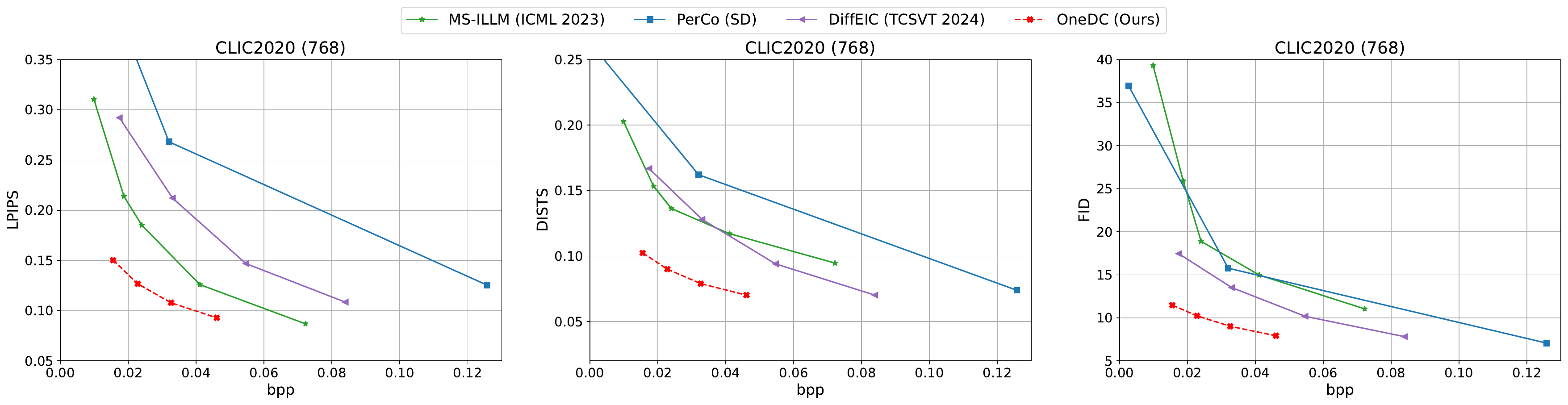}
    \vspace{-5mm}
    \caption{Rate–distortion curves on the CLIC2020 at \textit{resize \& center-crop, 256 FID} setting, with resized resolution 768.}
    \label{fig:supp_exp2}
    \vspace{-1mm}
\end{figure}

\textbf{Additional datasets and metrics.}
To enable comprehensive comparison, we further evaluate our method on the DIV2K test set~\cite{div2k} under the \textit{full-resolution} setting, as shown in Fig.~\ref{fig:supp_exp1}.
We also report results on the CLIC2020 test set at 768 resolution under DiffEIC’s protocol, i.e., \textit{resize \& center-crop, 256 FID}, as illustrated in Fig.~\ref{fig:supp_exp2}.
Across datasets, metrics, and evaluation settings, the OneDC consistently achieves SOTA performance, demonstrating strong robustness and generalization.
The raw data used for evaluation is available at Table.~\ref{tab:supp_3} $\sim$ ~\ref{tab:supp_8}.

\begin{figure}[t]
    \centering
    \includegraphics[width=\linewidth]{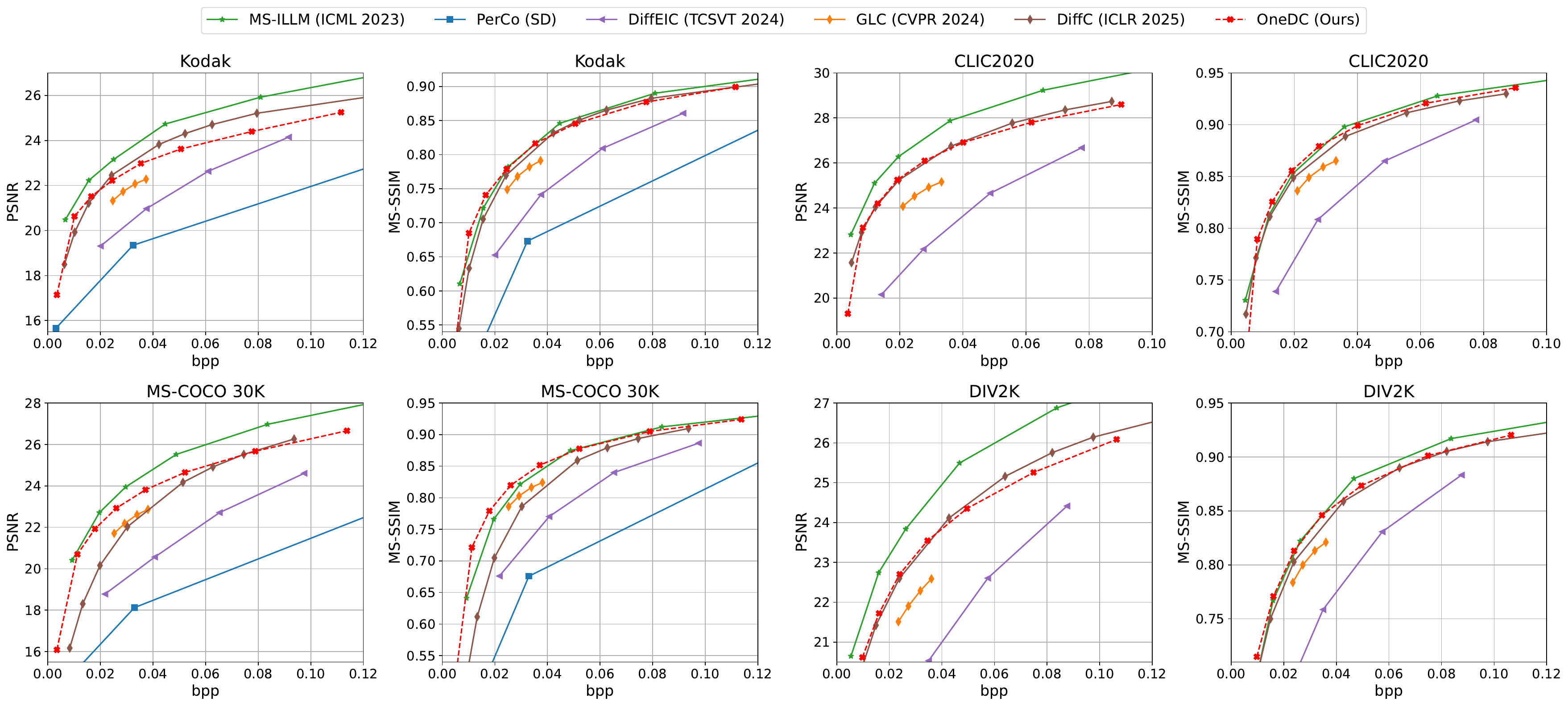}
    \vspace{-5mm}
    \caption{Comparison of methods measured by PSNR and MS-SSIM at \textit{full-resolution} setting.}
    \label{fig:supp_exp3}
    \vspace{-2mm}
\end{figure}

We also report traditional pixel-level distortion metrics (PSNR and MS-SSIM~\cite{wang2003multiscale}) to provide a more comprehensive analysis, as shown in Fig.~\ref{fig:supp_exp3}.
At extremely low bitrates, optimizing for PSNR often suppresses high-frequency details, resulting in blurred reconstructions~\cite{mentzer2020high}. 
While OneDC shows slightly lower PSNR than MS-ILLM, both perceptual metrics and qualitative examples clearly demonstrate its superior visual quality.
Despite prioritizing perceptual quality, OneDC still achieves competitive PSNR compared to other diffusion-based methods. 
Moreover, on the MS-SSIM metric, OneDC matches MS-ILLM and outperforms all other baselines.
These results confirm that OneDC delivers strong pixel-level fidelity alongside high perceptual realism, highlighting the overall effectiveness of our method.

\textbf{Additional visual examples.}
We present more qualitative comparisons across four datasets: Kodak (Fig.\ref{fig:supp_kodak}), CLIC2020—both full resolution (Figs.\ref{fig:supp_clic_1}) and $768 \times 768$ cropped (Fig.\ref{fig:supp_clic768}), and MS-COCO 30K (Fig.\ref{fig:supp_coco}).
OneDC consistently outperforms prior SOTA methods, delivering superior visual quality across diverse content and resolutions, yet with the lowest bitrate cost.

\textbf{Effectiveness of semantic distillation.}
To further evaluate the proposed semantic distillation strategy, we fine-tune the pretrained text-to-image one-step diffusion model \cite{yin2024improved}, replacing its textual semantic condition with the hyperprior features produced by our semantic hyperprior decoder $h_{sem}$.
This enables reconstruction using only the hyperprior signal.
Fig.~\ref{fig:supp_semantic} presents additional reconstruction results on the COCO2017 validation dataset \cite{lin2014microsoft}, extending Fig. 2 from the main paper.
These results confirm that the distilled model captures richer semantic information, facilitating final reconstruction.

\textbf{Bitrate allocation between hyperprior $\hat{z}$ and latent $\hat{z}$.} 
We ablate the roles of hyperprior $\hat{z}$ and latent $\hat{y}$ on CLIC2020 test set by fixing the hyperprior budget to 0.0035 bpp and gradually increasing the bitrate of $\hat{y}$.
Table~\ref{tab:supp_2} reports the results and reveals a clear division of each part’s role: 
(a) with zero bits allocated for latent $\hat{y}$, the codec still produces coherent reconstructions (also show in Fig.~\ref{fig:supp_semantic}), indicating that $\hat{z}$ supplies a strong semantic information; 
(b) as bits are assigned to $\hat{y}$, fidelity and perceptual quality improve monotonically (e.g., LPIPS drops from 0.290 to 0.119), confirming that $\hat{y}$ is essential for encoding fine-grained details.

\begin{table}[]
\centering
\caption{Complexity analysis with model size. BD-Rate is calculated on the MS-COCO 30K dataset.}
\label{tab:supp_1}
\vspace{1mm}
\resizebox{0.85\textwidth}{!}{
\begin{tabular}{@{}lllllll@{}}
\toprule
\multirow{2}{*}{Model} & \multirow{2}{*}{Params} & \multirow{2}{*}{Enc. Time (s)} & \multirow{2}{*}{Dec. Time (s)} & \multicolumn{3}{c}{BD-Rate (Metircs)↓} \\
 &  &  &  & LPIPS & DISTS & FID \\ \midrule
MS-ILLM & 181M & 0.14 & 0.17 & 138.3\% & 253.0\% & 478.4\% \\
DiffEIC & 1.4B & 0.32 & 12.4 & 305.0\% & 239.1\% & 341.0\% \\
PerCo (SD) & 3.8B+340M+955M* & 0.58 & 8.80 & 538.8\% & 345.8\% & 59.6\% \\
DiffC & 950M & 3.9$\sim$15.6 & 6.9$\sim$10.8 & 234.0\% & 196.1\% & 690.9\% \\
DDCM & 950M & - & - & - & - & - \\
OneDC & 1.4B & 0.15 & 0.34 & 0.00\% & 0.00\% & 0.00\% \\ \bottomrule
\multicolumn{7}{l}{\small \textcolor{gray}{*} Open-sourced PerCo includes an additional 3.8B BLIP2 caption model and 340M CLIP model.}
\end{tabular}
}
\end{table}

\begin{table}[]
\centering
\caption{Bitrate allocation analysis on the CLIC2020 dataset.}
\label{tab:supp_2}
\vspace{1mm}
\resizebox{0.85\textwidth}{!}{
\begin{tabular}{@{}llllllll@{}}
\toprule
Bpp $\hat{z}$ (ratio) & Bpp $\hat{y}$ (ratio) & Bpp Total & PSNR↑ & MS-SSIM↑ & LPIPS↓ & DISTS↓ & FID↓ \\ \midrule
0.0035 (100\%) & 0.0 (0\%) & 0.0035 & 19.31 & 0.629 & 0.290 & 0.169 & 14.885 \\
0.0035 (43\%) & 0.0047 (57\%) & 0.0082 & 23.13 & 0.790 & 0.163 & 0.089 & 6.223 \\
0.0035 (27\%) & 0.0094 (73\%) & 0.0129 & 24.20 & 0.826 & 0.139 & 0.077 & 5.560 \\
0.0035 (18\%) & 0.0157 (82\%) & 0.0192 & 25.25 & 0.856 & 0.119 & 0.068 & 4.979 \\ \bottomrule
\end{tabular}
}
\end{table}

\textbf{Model size.} We provide parameter counts, runtime comparisons (on 1024×1024 images), and BD-Rate results on the MS-COCO 30K dataset for better efficiency evaluation, as shown in Table~\ref{tab:supp_1}.
Compared to MS-ILLM, diffusion-based methods typically use larger models but achieve superior perceptual
quality (e.g., lower BD-Rate with FID) due to stronger generative capacity. Unlike other diffusion-based codecs,
OneDC avoids external caption models and multi-step sampling, enabling over 20× faster decoding while also achieving better rate-distortion performance.

\textbf{Memroy usage.} We also report memory usage under the \textit{resize \& center-crop} setting on $512 \times 512$ Kodak images.
PerCo (SD) requires about 22,220 MB of GPU memory, whereas DDCM uses 4,186 MB and OneDC uses 8,038 MB.
These results highlight that incorporating a large language model, as in PerCo, substantially increases computational burden.
DDCM achieves lower memory consumption by employing the SD model in a zero-shot manner.
Although OneDC requires more memory than DDCM, its single-step design reduces inference cost compared to both multi-step PerCo and DDCM.

\section{Training Details}
\textbf{Stage I training.}
This stage focuses on training the compression module and fine-tuning the one-step diffusion model \cite{yin2024improved} for the image reconstruction task.
The training loss is defined as:

\vspace{-3mm}
\begin{equation}
    L_{stageI} = L_{recon} + \lambda R + \alpha L_{aux}, \;\;\;\; \text{where} \;\; L_{recon}=L_{1}(x,\hat{x})+L_{perceptual}(x,\hat{x})
\end{equation}
\vspace{-4mm}

We use the L1 as the pixel-level loss and the LPIPS \cite{johnson2016perceptual} as the perceptual-level loss.
To support various bitrates, the rate-distortion trade-off parameter $\lambda$ is set to $\{0.6, 1.0, 1.8, 2.9, 4.6, 7.4, 12.2\}$. 
An auxiliary code prediction loss $L_{aux}$ is included with a weighting factor of $\alpha = 0.001$. We train our model on the dataset introduced in~\cite{gokaslan2023commoncanvas}.
Training is performed on 4×A100 GPUs for 800,000 steps, using a three-stage learning rate schedule with AdamW~\cite{loshchilov2017decoupled}:
a) 5e-5 for the first 500,000 steps; b) 1e-5 for the next 200,000 steps; c) 1e-6 for the final 100,000 steps.
During training, image patches of size $\{512, 1024\}$ are randomly cropped with probabilities of $\{0.6, 0.4\}$, respectively.
The batch size is set to 32 for $512 \times 512$ crops and 8 for $1024 \times 1024$ crops (across 4 GPUs).
This stage takes approximately 6 days as we use high-resolution patches for training.

\textbf{Stage II training.} 
This stage fine-tunes the one-step generator to improve reconstruction realism by better aligning the distribution of its outputs with that of real images. 
The training objective is:

\vspace{-5mm}
\begin{gather}
    L_{stageII} = L_{distill} + \beta L_{recon} + \gamma L_{adv}, \;\;\;\; \text{where:} \\
    L_{distill} = \mathbb{E}_{t,\tilde{y}'_{t}}[\epsilon_{fake}(\tilde{y}'_{t},t)-\epsilon_{real}(\tilde{y}'_{t},t)], \;\;  L_{adv} = \mathbb{E}_{t,\tilde{y}'_{t}}[-Disc(\epsilon_{fake}(\tilde{y}'_{t},t), t)]
\end{gather}
\vspace{-4mm}

Here, $L_{distill}$ represents the diffusion distillation loss \cite{yin2024one}, and $L_{adv}$  is the adversarial loss in the latent space, following \cite{yin2024improved}.
$Disc$ denotes the discriminator network, which takes the mid feature in the diffusion U-Net as the input \cite{yin2024improved}.
The variable $\tilde{y}_0$ is the latent output generated by the one-step diffusion model, and $\tilde{y}'_t$ is its noised version at timestep $t$.
We uniformly sample $t \in [20, 640]$, since synthesizing high-frequency details does not require large noise levels.
The weighting parameters are set as follows: $\beta = 0.625$ balance the reconstruction and distillation terms, and $\gamma=0.001$ for the adversarial loss ($\gamma$ follows \cite{yin2024improved}).

Specifically, diffusion distillation \cite{yin2024one} minimizes the expected Kullback-Leibler (KL) divergence between the time-dependent distributions of the target $p_{real, t}$ and the generator output $p_{real, t}$, thereby effectively transferring knowledge from the multi-step diffusion model to the one-step generator.
The gradient used to update the one-step generator parameters $\theta$ is given by the difference between the score functions of the real and fake distributions:

\vspace{-5mm}
\begin{align}
    \nabla_{\theta} L_{distill} &= \mathbb{E}_{t, \tilde{y}_0}(\nabla_{\theta}\text{KL}(p_{fake, t}||p_{real, t})) \\
    &=-\mathbb{E}_{t, \tilde{y}'_t}[(s_{real}(\tilde{y}'_t, t)-s_{fake}(\tilde{y}'_t, t))\frac{d\epsilon_{\theta}}{d\theta}] \\
    &=\mathbb{E}_{t, \tilde{y}'_t}[(\epsilon_{fake}(\tilde{y}'_t, t)-\epsilon_{real}(\tilde{y}'_t, t))\frac{d\epsilon_{\theta}}{d\theta}]
\end{align}
\vspace{-5mm}

Here, $s_{\text{real}}$ and $s_{\text{fake}}$ are the score functions learned by multi-step diffusion model $\epsilon_{\text{real}}$ and $\epsilon_{\text{fake}}$ respectively.
To ensure the fake score network $\epsilon_{\text{fake}}$ accurately tracks the evolving distribution of the one-step diffusion model, we update it using a standard denoising loss:

\vspace{-3mm}
\begin{equation}
    L_{fake} = \mathbb{E}_{t, \tilde{y}'_t}||\epsilon_{fake}(\tilde{y}'_t,t)-\tilde{y}_{0}||^{2}_{2}
\end{equation}
\vspace{-5mm}

The improved version of diffusion distillation proposed in \cite{yin2024improved} introduces adversarial training in the latent space to further enhance distribution alignment. 
A discriminator is trained to differentiate between features extracted from real and generated images, using the following objective:

\vspace{-3mm}
\begin{equation}
    L_{Disc} = \mathbb{E}_{t, x}[Disc(\epsilon_{fake}(\tilde{y}'_t, t)) - Disc(\epsilon_{fake}(E_\text{VAE}(x)+n_{t}, t))]
\end{equation}
\vspace{-5mm}

Here, $E_{\text{VAE}}$ denotes the encoder of the diffusion model, and $n_{t}$ is the sampled noise in timestep $t$.
The fake branch is then optimized with an additional adversarial loss:

\vspace{-3mm}
\begin{equation}
    L = L_{fake} + \sigma L_{Disc}
\end{equation}
\vspace{-5mm}

We follow \cite{yin2024improved} and set $\sigma=0.01$.
During training, the fake branch and discriminator are each updated 10 times for every update of the one-step generator, ensuring stable adversarial optimization.

Training is conducted on 4×A100 GPUs for 1,000,000 steps.
This stage requires around 12 days, as diffusion
distillation introduces additional cost in addition to high-resolution training.
The learning rate is fixed at 1e-6 (with AdamW) for the one-step generator, fake network, and discriminator.
The batch size, cropping strategy and training data are identical to those used in Stage I.

\section{Model Details}
\begin{figure}[t]
    \centering
    \includegraphics[width=\linewidth]{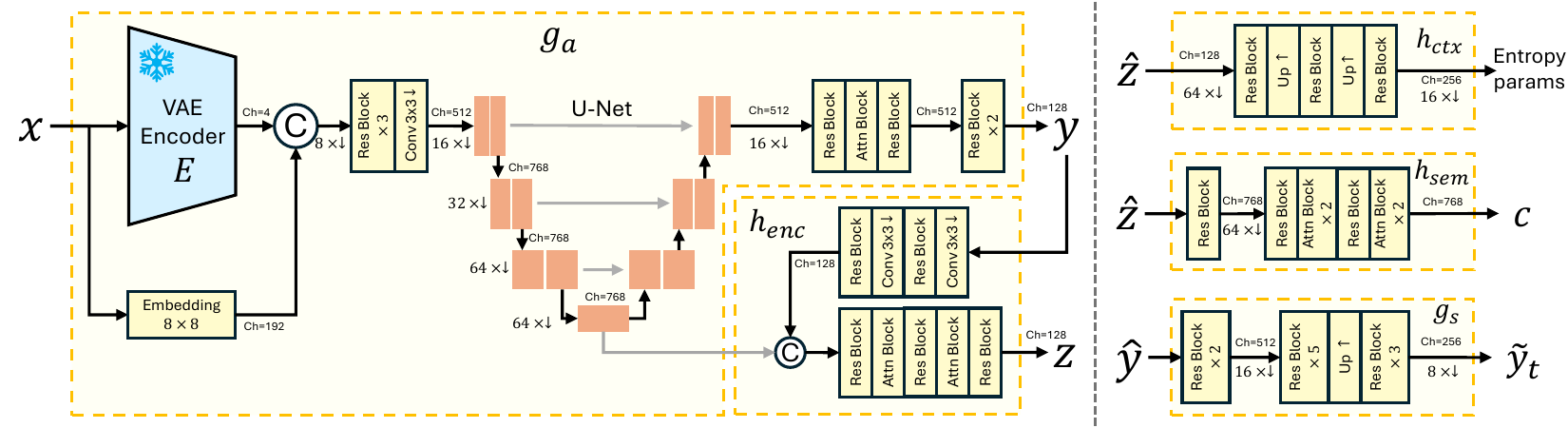}
    \vspace{-5mm}
    \caption{Details of our model architecture. The pratrained VAE encoder is from SD 1.5. For the U-Net used in $g_a$, we use the implementation from the diffusers library \cite{diffusers}.}
    \label{fig:model1}
    \vspace{-2mm}
\end{figure}

The overall architecture is shown in Fig.\ref{fig:model1}.
To ensure better alignment with the latent space of the diffusion model, we extract features from its pretrained VAE encoder.
Additionally, inspired by DiffEIC\cite{li2024towards}, we incorporate embeddings from the original input image to enrich the encoder with complementary spatial and semantic information.
To jointly learn compressible latents and capture high-level semantics, we introduce a lightweight U-Net within the analysis transform.
Its multi-scale design facilitates effective aggregation of both local textures and global structures.
The extracted semantic features are integrated into the hyperprior branch, enhancing its representational capability.

For parameter-efficient adaptation, we insert LoRA \cite{hu2022lora} layers across all modules of the pretrained one-step diffusion U-Net, setting the LoRA rank to 64. 
The combined parameter count of the encoder and decoder components ($g_a, h_{enc}, h_{ctx}, h_{sem}$ and $g_s$) is 394M, while the adapted diffusion U-Net contributes an additional 928M parameters (about 860M for SD1.5 and 68M for LoRA).

\begin{figure}[t]
    \centering
    \includegraphics[width=0.5\linewidth]{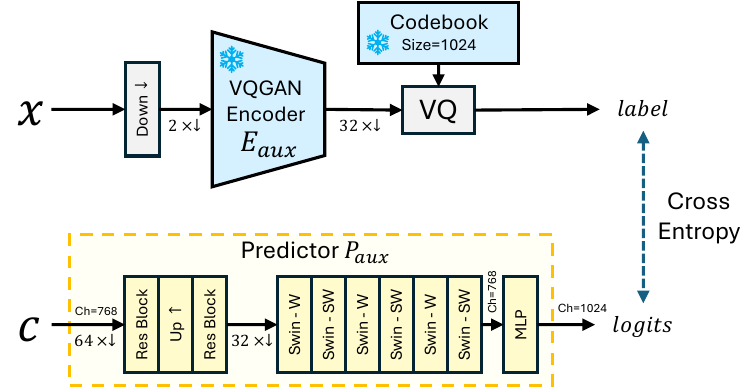}
    \vspace{-1mm}
    \caption{Details of the auxiliary code predictor module. For Swin Transformer block \cite{liu2021swin}, we set window size to 16, head dim to 64. W means normal window, SW means shifted window.}
    \label{fig:model2}
    \vspace{-2mm}
\end{figure}

Further details of the semantic distillation mechanism are provided in Fig.~\ref{fig:model2}.
We adopt a Swin Transformer \cite{liu2021swin} to improve adaptability across different image resolutions during training. 

\section{Social Impact}
\textbf{Positive Aspects.}
Generative codecs offer substantial benefits by significantly reducing the storage and transmission demands of high-resolution imagery through the synthesis of perceptually convincing content. 
This improvement in efficiency can help broaden access to high-quality visual media, particularly in bandwidth-limited or resource-constrained environments. 
The resulting gains in data economy and speed of delivery have promising implications for applications in social communication and personal entertainment.

\textbf{Negative Aspects.}
Despite these advantages, generative compression introduces synthesized content that may deviate from the original input, raising concerns about the fidelity and authenticity of reconstructed images. 
Ongoing research into multi-realism codecs~\cite{agustsson2023multi} offers a potential path forward, enabling more transparent control over the trade-off between realism and fidelity.

%% file: section_7_append_figures.tex
\begin{figure}
    \centering
    \includegraphics[width=1.0\linewidth]{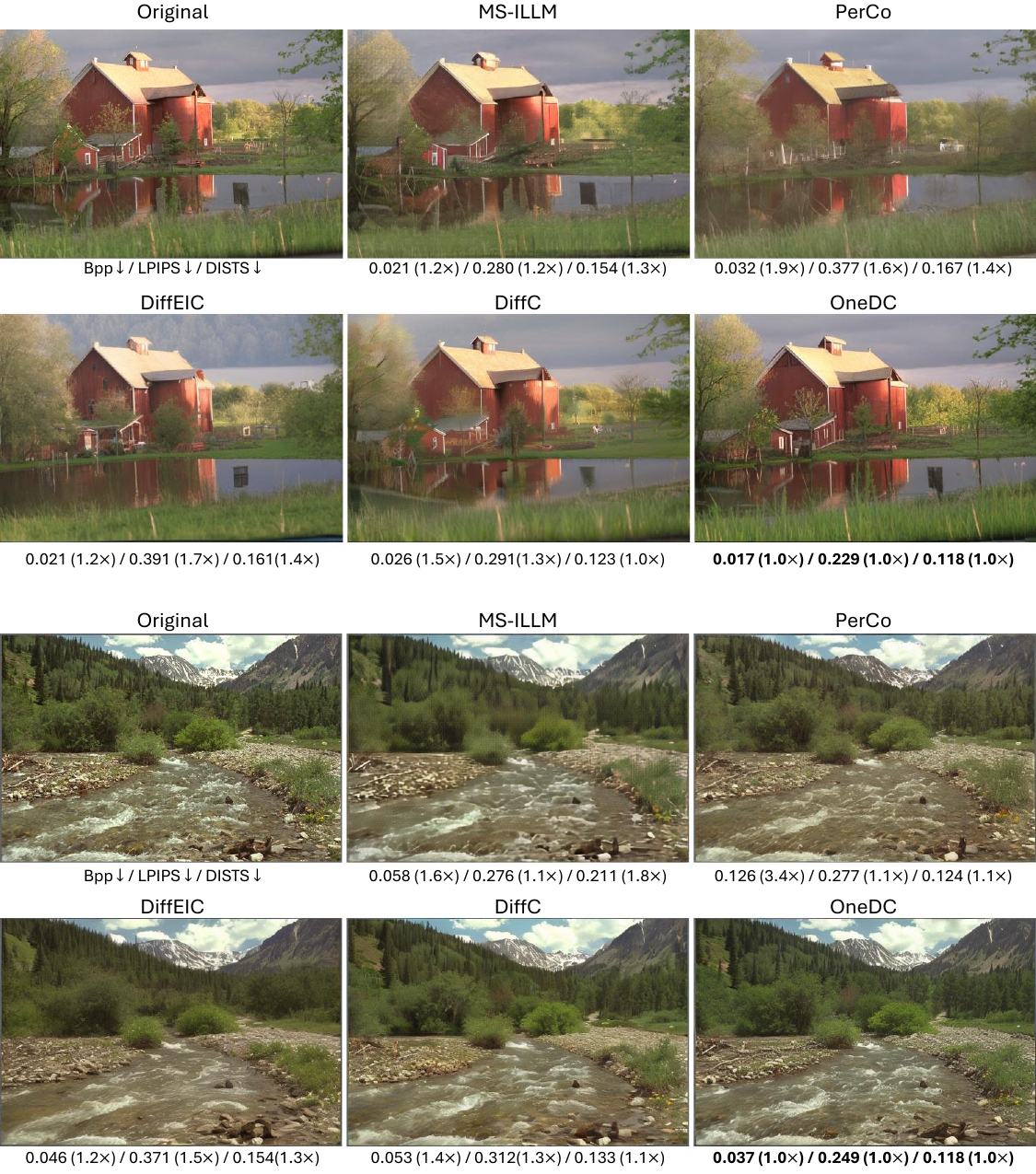}
    \caption{Visual results on the Kodak dataset. 
    The VAE-based MS-ILLM exhibits noticeable artifacts. Compared with our OneDC, the previous SOTA multi-step diffusion codec DiffC requires at least $1.4 \times$ bitrate on these two examples.
    Zoom in for better view.
    }
    \label{fig:supp_kodak}
\end{figure}

\begin{figure}
    \centering
    \includegraphics[width=1.0\linewidth]{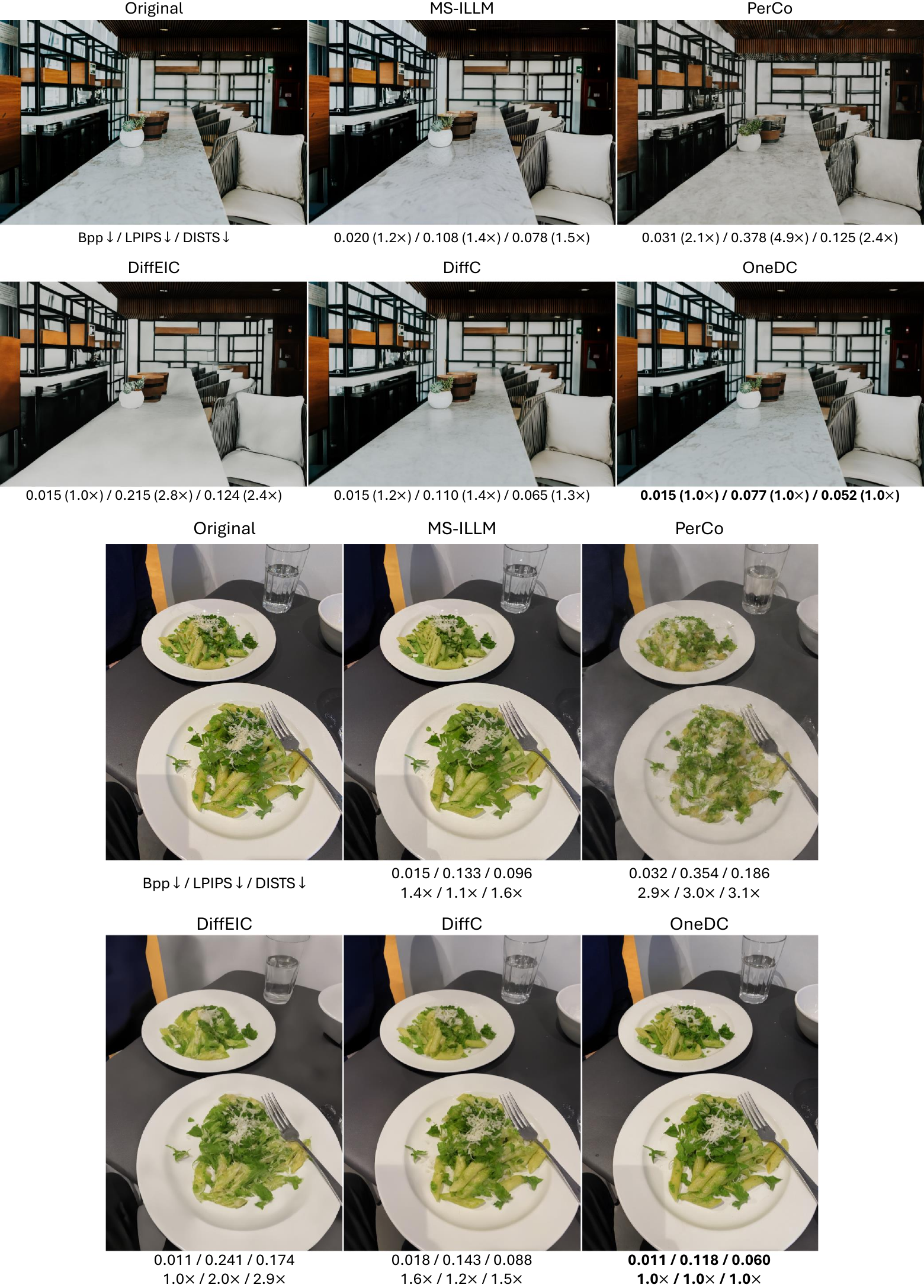}
    \caption{Visual results on the CLIC2020 test set. 
    Zoom in for better view. }
    \label{fig:supp_clic_1}
\end{figure}


\begin{figure}
    \centering
    \includegraphics[width=1.0\linewidth]{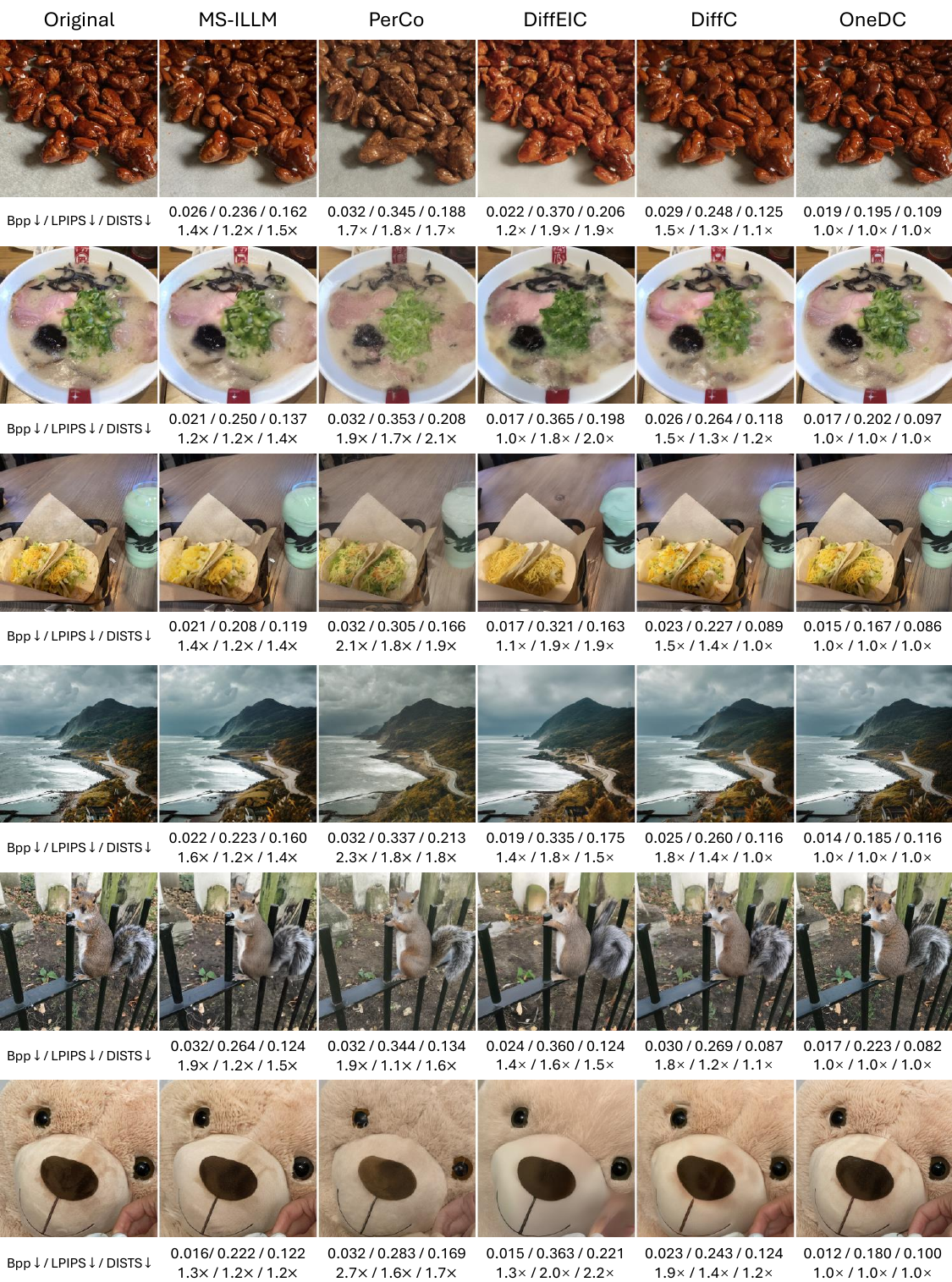}
    \caption{Visual results on the CLIC2020 test set ($768 \times 768$). Compared with OneDC, the previous SOTA multi-step diffusion method DiffC requires $\geq1.5$ higher bitrate while still producing slightly lower reconstruction quality on these examples.
    Zoom in for better view.}
    \label{fig:supp_clic768}
\end{figure}

\begin{figure}
    \centering
    \includegraphics[width=1.0\linewidth]{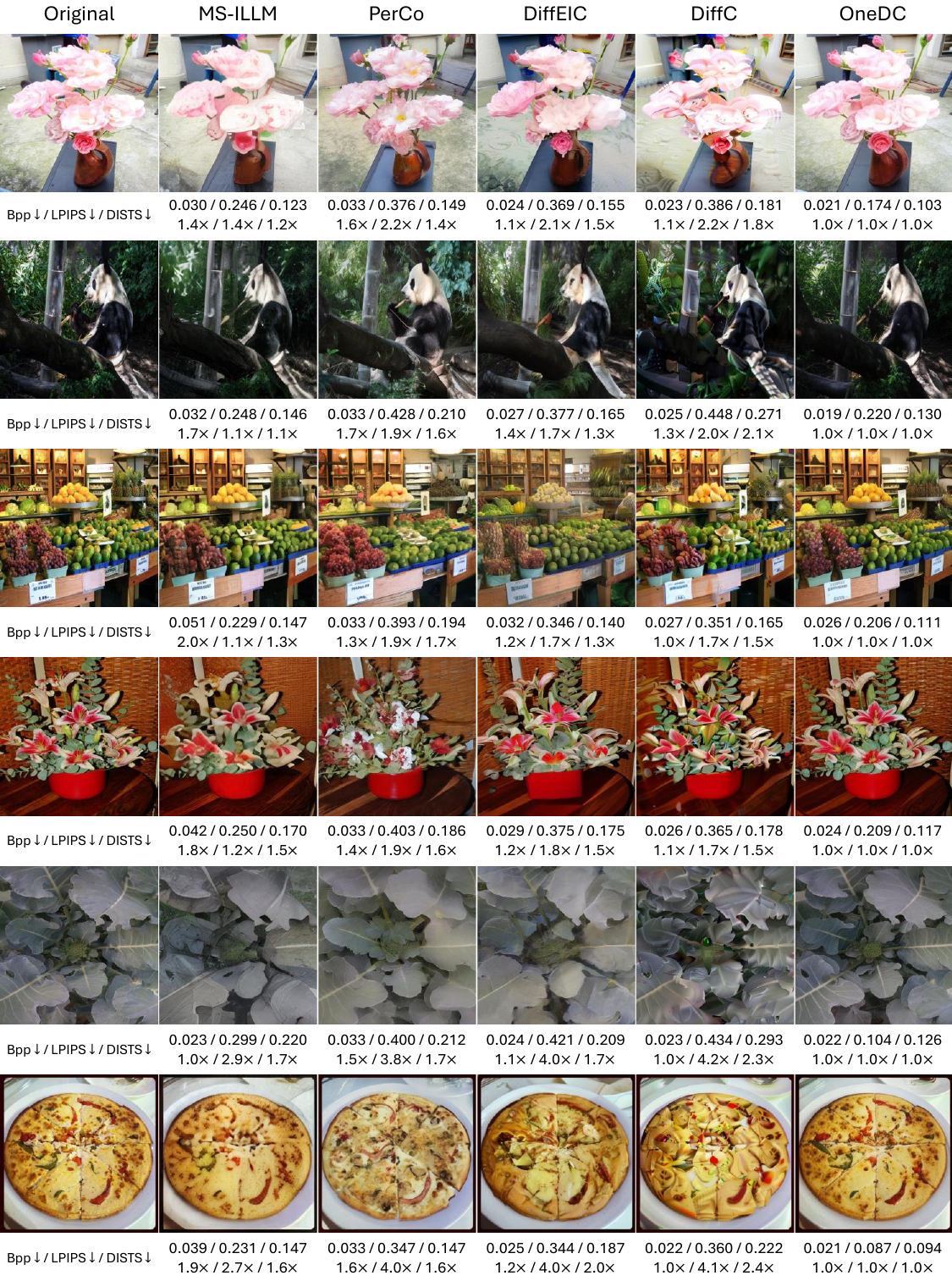}
    \caption{Visual results on the MS-COCO 30K dataset. PerCo, the previous best-performing method in terms of FID, requires $\geq1.4$  bitrate over OneDC, but still results in suboptimal fidelity.
    Zoom in for better view.}
    \label{fig:supp_coco}
\end{figure}

\begin{figure}
    \centering
    \includegraphics[width=0.7\linewidth]{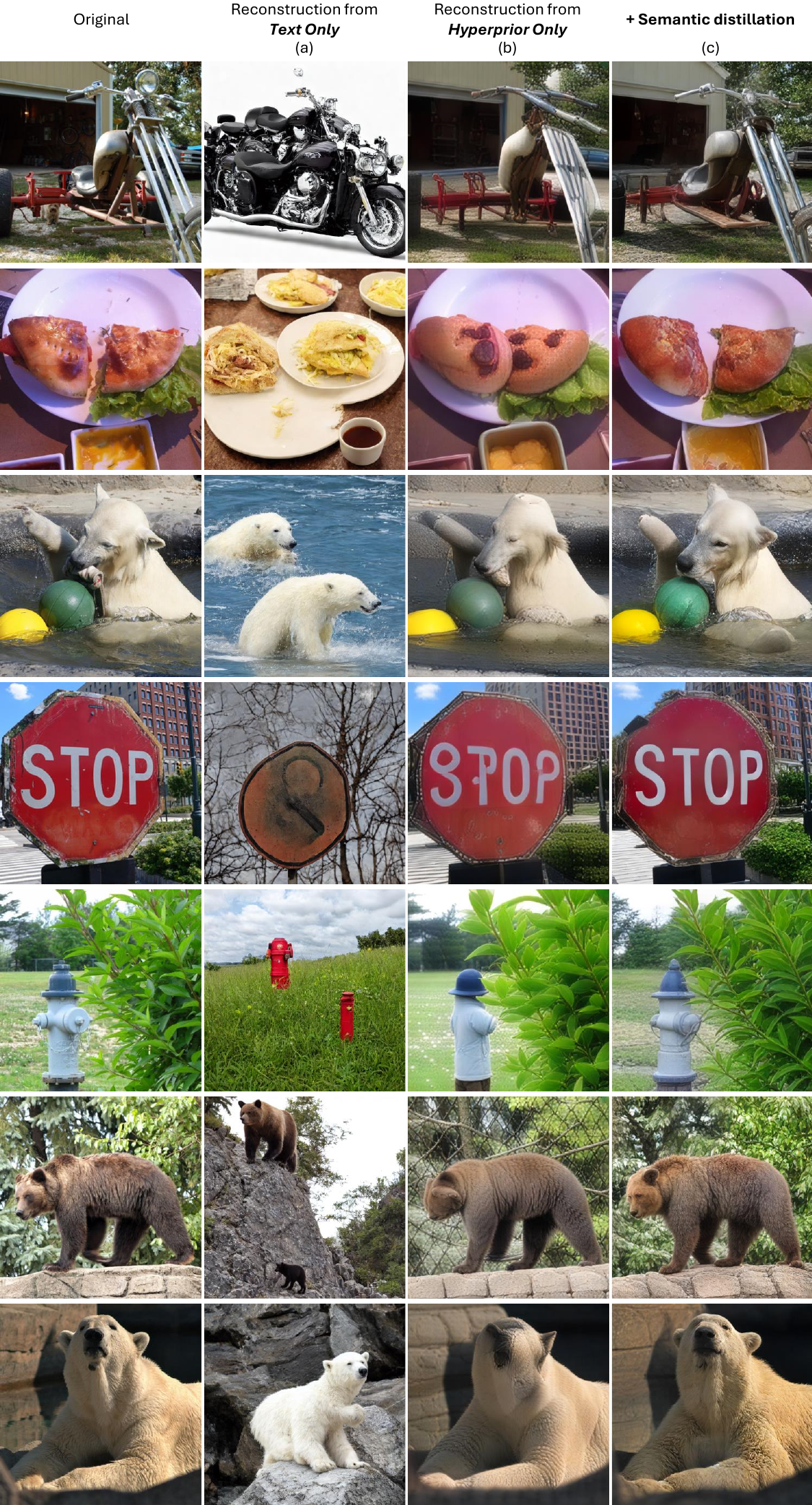}
    \caption{Reconstructions from different semantic guidance. (a) Text prompts struggle to capture complex visual semantics, resulting in severe distortions when using a pretrained text-to-image one-step diffusion model~\cite{yin2024improved}. (b) We finetune the model \cite{yin2024improved} for hyperprior-to-image generation. Hyperprior guidance yields more faithful reconstructions. (c) Our proposed semantic distillation further improves object-level accuracy.}
    \label{fig:supp_semantic}
\end{figure}

%% file: section_8_absolute_number.tex
\begin{table}[]
\centering
\caption{Compression quantitative evaluations, for the Kodak dataset with full-resolution.}
\label{tab:supp_3}
\vspace{1mm}
\resizebox{0.7\textwidth}{!}{
\begin{tabular}{@{}llllll@{}}
\toprule
Method & BPP & PSNR & MS-SSIM & LPIPS & DISTS \\ \midrule
MS-ILLM & 0.0066 & 20.478 & 0.611 & 0.438 & 0.280 \\
 & 0.0156 & 22.224 & 0.722 & 0.292 & 0.177 \\
 & 0.0250 & 23.155 & 0.782 & 0.228 & 0.147 \\
 & 0.0447 & 24.731 & 0.846 & 0.158 & 0.135 \\
 & 0.0809 & 25.922 & 0.890 & 0.110 & 0.109 \\
 & 0.1535 & 27.532 & 0.928 & 0.073 & 0.081 \\
 & 0.2962 & 29.634 & 0.960 & 0.045 & 0.061 \\ \midrule
PerCo (SD) & 0.0031 & 15.662 & 0.418 & 0.529 & 0.234 \\
 & 0.0324 & 19.344 & 0.673 & 0.299 & 0.162 \\
 & 0.1261 & 22.967 & 0.847 & 0.141 & 0.080 \\ \midrule
DiffEIC & 0.0200 & 19.308 & 0.653 & 0.326 & 0.169 \\
 & 0.0375 & 20.970 & 0.741 & 0.242 & 0.134 \\
 & 0.0610 & 22.631 & 0.809 & 0.173 & 0.098 \\
 & 0.0916 & 24.147 & 0.861 & 0.129 & 0.077 \\ \midrule
GLC & 0.0247 & 21.320 & 0.749 & 0.196 & 0.113 \\
 & 0.0286 & 21.729 & 0.768 & 0.180 & 0.104 \\
 & 0.0331 & 22.071 & 0.782 & 0.168 & 0.098 \\
 & 0.0374 & 22.279 & 0.791 & 0.161 & 0.095 \\ \midrule
DiffC & 0.0063 & 18.494 & 0.545 & 0.450 & 0.201 \\
 & 0.0102 & 19.926 & 0.633 & 0.363 & 0.165 \\
 & 0.0155 & 21.212 & 0.705 & 0.288 & 0.139 \\
 & 0.0242 & 22.453 & 0.770 & 0.225 & 0.112 \\
 & 0.0423 & 23.823 & 0.832 & 0.163 & 0.087 \\
 & 0.0522 & 24.304 & 0.851 & 0.144 & 0.080 \\
 & 0.0625 & 24.704 & 0.865 & 0.130 & 0.074 \\
 & 0.0795 & 25.207 & 0.883 & 0.113 & 0.065 \\
 & 0.1227 & 25.949 & 0.905 & 0.091 & 0.055 \\ \midrule
OneDC & 0.0034 & 17.141 & 0.472 & 0.380 & 0.204 \\
 & 0.0101 & 20.631 & 0.685 & 0.220 & 0.131 \\
 & 0.0165 & 21.513 & 0.741 & 0.183 & 0.110 \\
 & 0.0245 & 22.220 & 0.779 & 0.154 & 0.097 \\
 & 0.0354 & 22.979 & 0.817 & 0.133 & 0.085 \\
 & 0.0506 & 23.620 & 0.845 & 0.114 & 0.076 \\
 & 0.0775 & 24.396 & 0.877 & 0.096 & 0.066 \\
 & 0.1115 & 25.248 & 0.899 & 0.083 & 0.057 \\ \bottomrule
\end{tabular}
}
\end{table}

\clearpage

\begin{table}[]
\centering
\caption{Compression quantitative evaluations, for the CLIC2020 test set with full-resolution.}
\label{tab:supp_4}
\vspace{1mm}
\resizebox{0.8\textwidth}{!}{
\begin{tabular}{@{}lllllll@{}}
\toprule
Method & BPP & PSNR & MS-SSIM & LPIPS & DISTS & FID \\ \midrule
MS-ILLM & 0.0045 & 22.815 & 0.731 & 0.365 & 0.216 & 44.954 \\
 & 0.0120 & 25.100 & 0.813 & 0.225 & 0.131 & 15.329 \\
 & 0.0195 & 26.275 & 0.853 & 0.173 & 0.106 & 8.748 \\
 & 0.0359 & 27.875 & 0.898 & 0.114 & 0.086 & 6.261 \\
 & 0.0654 & 29.229 & 0.928 & 0.080 & 0.066 & 4.481 \\
 & 0.1240 & 30.865 & 0.953 & 0.054 & 0.048 & 2.645 \\
 & 0.2377 & 32.833 & 0.972 & 0.035 & 0.035 & 1.660 \\ \midrule
PerCo (SD) & 0.0022 & 15.340 & 0.473 & 0.589 & 0.326 & 76.658 \\
 & 0.0315 & 16.733 & 0.537 & 0.389 & 0.194 & 18.039 \\
 & 0.1249 & 17.419 & 0.554 & 0.306 & 0.091 & 5.013 \\ \midrule
DiffEIC & 0.0142 & 20.152 & 0.739 & 0.308 & 0.185 & 18.427 \\
 & 0.0275 & 22.173 & 0.808 & 0.222 & 0.132 & 11.039 \\
 & 0.0487 & 24.656 & 0.865 & 0.153 & 0.089 & 7.278 \\
 & 0.0776 & 26.673 & 0.905 & 0.109 & 0.060 & 4.942 \\ \midrule
DiffC & 0.0047 & 21.571 & 0.717 & 0.311 & 0.179 & 20.076 \\
 & 0.0079 & 22.914 & 0.771 & 0.253 & 0.144 & 12.332 \\
 & 0.0123 & 24.040 & 0.811 & 0.210 & 0.117 & 8.957 \\
 & 0.0197 & 25.241 & 0.849 & 0.171 & 0.092 & 6.813 \\
 & 0.0362 & 26.745 & 0.889 & 0.128 & 0.067 & 5.252 \\
 & 0.0556 & 27.772 & 0.911 & 0.103 & 0.053 & 4.414 \\
 & 0.0724 & 28.357 & 0.923 & 0.090 & 0.046 & 3.929 \\
 & 0.0872 & 28.733 & 0.930 & 0.082 & 0.042 & 3.685 \\ \midrule
OneDC & 0.0035 & 19.305 & 0.629 & 0.290 & 0.169 & 14.885 \\
 & 0.0083 & 23.129 & 0.789 & 0.164 & 0.089 & 6.223 \\
 & 0.0130 & 24.201 & 0.826 & 0.139 & 0.077 & 5.560 \\
 & 0.0192 & 25.248 & 0.856 & 0.119 & 0.068 & 4.979 \\
 & 0.0279 & 26.096 & 0.879 & 0.103 & 0.060 & 4.234 \\
 & 0.0401 & 26.912 & 0.899 & 0.090 & 0.053 & 3.581 \\
 & 0.0617 & 27.804 & 0.921 & 0.076 & 0.044 & 2.977 \\
 & 0.0902 & 28.596 & 0.936 & 0.065 & 0.037 & 2.410 \\ \bottomrule
\end{tabular}
}
\end{table}

\clearpage

\begin{table}[]
\centering
\caption{Compression quantitative evaluations, for the MS-COCO 30K dataset at $512\times512$ resolution.}
\label{tab:supp_5}
\vspace{1mm}
\resizebox{0.8\textwidth}{!}{
\begin{tabular}{@{}lllllll@{}}
\toprule
Method & BPP & PSNR & MS-SSIM & LPIPS & DISTS & FID \\ \midrule
MS-ILLM & 0.0092 & 20.411 & 0.641 & 0.397 & 0.255 & 72.693 \\
 & 0.0196 & 22.708 & 0.766 & 0.257 & 0.169 & 17.992 \\
 & 0.0296 & 23.948 & 0.821 & 0.200 & 0.145 & 9.041 \\
 & 0.0488 & 25.517 & 0.875 & 0.138 & 0.122 & 4.100 \\
 & 0.0835 & 26.972 & 0.912 & 0.095 & 0.099 & 2.032 \\
 & 0.1510 & 28.734 & 0.944 & 0.063 & 0.075 & 0.990 \\
 & 0.2850 & 30.930 & 0.969 & 0.039 & 0.058 & 0.457 \\ \midrule
PerCo (SD) & 0.0036 & 14.126 & 0.388 & 0.545 & 0.245 & 4.467 \\
 & 0.0329 & 18.124 & 0.676 & 0.311 & 0.159 & 2.748 \\
 & 0.1267 & 22.802 & 0.869 & 0.134 & 0.080 & 1.152 \\ \midrule
DiffEIC & 0.0217 & 18.778 & 0.676 & 0.318 & 0.171 & 6.151 \\
 & 0.0407 & 20.564 & 0.770 & 0.229 & 0.133 & 3.929 \\
 & 0.0653 & 22.710 & 0.840 & 0.159 & 0.099 & 2.578 \\
 & 0.0975 & 24.606 & 0.886 & 0.116 & 0.077 & 1.911 \\ \midrule
DiffC & 0.0083 & 16.170 & 0.495 & 0.524 & 0.271 & 90.989 \\
 & 0.0133 & 18.306 & 0.611 & 0.427 & 0.229 & 57.542 \\
 & 0.0198 & 20.159 & 0.705 & 0.334 & 0.187 & 28.909 \\
 & 0.0303 & 22.029 & 0.786 & 0.241 & 0.140 & 9.763 \\
 & 0.0514 & 24.170 & 0.859 & 0.155 & 0.095 & 2.805 \\
 & 0.0628 & 24.915 & 0.879 & 0.132 & 0.082 & 2.135 \\
 & 0.0745 & 25.520 & 0.894 & 0.115 & 0.074 & 1.809 \\
 & 0.094 & 26.264 & 0.910 & 0.098 & 0.064 & 1.518 \\ \midrule
OneDC & 0.0034 & 16.083 & 0.460 & 0.376 & 0.211 & 13.607 \\
 & 0.0112 & 20.696 & 0.721 & 0.201 & 0.128 & 3.496 \\
 & 0.0179 & 21.924 & 0.779 & 0.166 & 0.111 & 2.817 \\
 & 0.0260 & 22.924 & 0.820 & 0.139 & 0.097 & 2.379 \\
 & 0.0371 & 23.809 & 0.852 & 0.118 & 0.086 & 2.044 \\
 & 0.0521 & 24.650 & 0.878 & 0.101 & 0.076 & 1.719 \\
 & 0.0789 & 25.674 & 0.905 & 0.082 & 0.064 & 1.365 \\
 & 0.1137 & 26.662 & 0.924 & 0.069 & 0.055 & 1.043 \\ \bottomrule
\end{tabular}
}
\end{table}

\clearpage

\begin{table}[]
\centering
\caption{Compression quantitative evaluations, for the DIV2K test set at full-resolution.}
\label{tab:supp_6}
\vspace{1mm}
\resizebox{0.8\textwidth}{!}{
\begin{tabular}{@{}lllllll@{}}
\toprule
Method & BPP & PSNR & MS-SSIM & LPIPS & DISTS & FID \\ \midrule
MS-ILLM & 0.0054 & 20.647 & 0.647 & 0.396 & 0.234 & 79.455 \\
 & 0.0159 & 22.741 & 0.767 & 0.254 & 0.143 & 33.303 \\
 & 0.0262 & 23.837 & 0.822 & 0.198 & 0.116 & 22.428 \\
 & 0.0467 & 25.493 & 0.880 & 0.138 & 0.102 & 18.537 \\
 & 0.0837 & 26.878 & 0.917 & 0.097 & 0.080 & 13.574 \\
 & 0.1547 & 28.503 & 0.946 & 0.065 & 0.055 & 8.070 \\
 & 0.2895 & 30.545 & 0.969 & 0.040 & 0.040 & 5.123 \\ \midrule
PerCo (SD) & 0.0022 & 14.529 & 0.381 & 0.570 & 0.320 & 95.345 \\
 & 0.0316 & 15.411 & 0.432 & 0.417 & 0.218 & 36.671 \\
 & 0.1249 & 15.340 & 0.405 & 0.355 & 0.105 & 11.314 \\ \midrule
DiffEIC & 0.0182 & 18.615 & 0.663 & 0.327 & 0.190 & 34.527 \\
 & 0.0349 & 20.532 & 0.759 & 0.241 & 0.132 & 21.656 \\
 & 0.0575 & 22.604 & 0.831 & 0.172 & 0.089 & 14.026 \\
 & 0.0876 & 24.415 & 0.883 & 0.125 & 0.061 & 9.644 \\ \midrule
DiffC & 0.0056 & 19.091 & 0.623 & 0.353 & 0.198 & 45.767 \\
 & 0.0096 & 20.345 & 0.695 & 0.291 & 0.158 & 32.086 \\
 & 0.0149 & 21.422 & 0.750 & 0.244 & 0.127 & 24.270 \\
 & 0.0238 & 22.599 & 0.803 & 0.199 & 0.098 & 18.541 \\
 & 0.0427 & 24.113 & 0.859 & 0.149 & 0.071 & 14.098 \\
 & 0.0640 & 25.158 & 0.890 & 0.120 & 0.056 & 11.859 \\
 & 0.0820 & 25.755 & 0.905 & 0.104 & 0.048 & 10.622 \\
 & 0.0976 & 26.143 & 0.914 & 0.096 & 0.044 & 9.938 \\
 & 0.1273 & 26.645 & 0.925 & 0.085 & 0.039 & 8.983 \\ \midrule
OneDC & 0.0097 & 20.615 & 0.715 & 0.200 & 0.103 & 15.874 \\
 & 0.0161 & 21.719 & 0.771 & 0.168 & 0.086 & 13.680 \\
 & 0.0239 & 22.703 & 0.813 & 0.144 & 0.077 & 12.365 \\
 & 0.0345 & 23.546 & 0.846 & 0.124 & 0.067 & 11.242 \\
 & 0.0495 & 24.351 & 0.873 & 0.108 & 0.060 & 10.096 \\
 & 0.0749 & 25.257 & 0.901 & 0.090 & 0.050 & 8.325 \\
 & 0.1064 & 26.086 & 0.920 & 0.077 & 0.040 & 6.822 \\ \bottomrule
\end{tabular}
}
\end{table}

\clearpage

\begin{table}[]
\centering
\caption{Compression quantitative evaluations, for the Kodak dataset at \textit{resize \& center-crop} setting with $512\times 512$ resolution.}
\label{tab:supp_7}
\vspace{1mm}
\resizebox{0.5\textwidth}{!}{
\begin{tabular}{@{}lllll@{}}
\toprule
Method & BPP & PSNR & LPIPS & FID \\ \midrule
MS-ILLM & 0.085 & 25.683 & 0.11 & 36.08 \\
 & 0.159 & 27.296 & 0.072 & 28.556 \\
 & 0.304 & 29.395 & 0.044 & 24.448 \\ \midrule
PerCo (SD) & 0.033 & 19.017 & 0.307 & 37.019 \\
 & 0.127 & 22.325 & 0.145 & 26.418 \\ \midrule
DDCM & 0.03 & 22.066 & 0.222 & 32.031 \\
 & 0.038 & 22.551 & 0.19 & 29.117 \\
 & 0.05 & 23.013 & 0.161 & 25.647 \\
 & 0.095 & 23.606 & 0.138 & 24.215 \\
 & 0.149 & 24.069 & 0.124 & 23.199 \\ \midrule
OneDC & 0.0109 & 20.259 & 0.223 & 36.862 \\
 & 0.0177 & 21.219 & 0.183 & 33.075 \\
 & 0.0257 & 21.970 & 0.154 & 30.947 \\
 & 0.0369 & 22.718 & 0.132 & 29.103 \\
 & 0.0524 & 23.363 & 0.113 & 27.428 \\
 & 0.0795 & 24.202 & 0.095 & 25.452 \\
 & 0.1139 & 25.021 & 0.082 & 23.448 \\ \bottomrule
\end{tabular}
}
\end{table}

\begin{table}[]
\centering
\caption{Compression quantitative evaluations, for the CLIC2020 test set at \textit{resize \& center-crop} setting with $768\times 768$ resolution.}
\label{tab:supp_8}
\vspace{1mm}
\resizebox{0.5\textwidth}{!}{
\begin{tabular}{@{}lllll@{}}
\toprule
Method & BPP & PSNR & LPIPS & FID \\ \midrule
MS-ILLM & 0.006 & 21.304 & 0.447 & 62.327 \\
 & 0.009 & 22.703 & 0.34 & 41.479 \\
 & 0.072 & 27.808 & 0.084 & 7.852 \\
 & 0.134 & 29.567 & 0.055 & 5.966 \\ \midrule
PerCo (SD) & 0.003 & 15.339 & 0.517 & 30.409 \\
 & 0.032 & 19.018 & 0.269 & 12.869 \\
 & 0.126 & 23.387 & 0.122 & 5.419 \\ \midrule
DDCM & 0.007 & 19.672 & 0.404 & 23.862 \\
 & 0.008 & 20.532 & 0.354 & 19.521 \\
 & 0.01 & 21.207 & 0.314 & 15.559 \\
 & 0.014 & 22.116 & 0.262 & 11.362 \\
 & 0.017 & 22.722 & 0.227 & 8.722 \\
 & 0.022 & 23.366 & 0.192 & 6.753 \\
 & 0.042 & 24.136 & 0.156 & 5.051 \\
 & 0.066 & 24.739 & 0.133 & 4.549 \\
 & 0.137 & 25.65 & 0.108 & 4.132 \\ \midrule
OneDC & 0.0034 & 17.619 & 0.326 & 15.325 \\
 & 0.0098 & 22.378 & 0.172 & 7.953 \\
 & 0.0155 & 23.600 & 0.140 & 7.390 \\
 & 0.0228 & 24.641 & 0.117 & 7.011 \\
 & 0.0326 & 25.532 & 0.099 & 6.538 \\
 & 0.0461 & 26.416 & 0.084 & 5.874 \\
 & 0.0694 & 27.463 & 0.069 & 4.915 \\
 & 0.0993 & 28.471 & 0.058 & 4.046 \\ \bottomrule
\end{tabular}
}
\end{table}

%% file: main.bbl
\begin{thebibliography}{72}
\providecommand{\natexlab}[1]{#1}
\providecommand{\url}[1]{\texttt{#1}}
\expandafter\ifx\csname urlstyle\endcsname\relax
  \providecommand{\doi}[1]{doi: #1}\else
  \providecommand{\doi}{doi: \begingroup \urlstyle{rm}\Url}\fi

\bibitem[Agustsson and Timofte(2017)]{div2k}
Eirikur Agustsson and Radu Timofte.
\newblock Ntire 2017 challenge on single image super-resolution: Dataset and study.
\newblock In \emph{The IEEE Conference on Computer Vision and Pattern Recognition (CVPR) Workshops}, July 2017.
\newblock URL \url{http://www.vision.ee.ethz.ch/~timofter/publications/Agustsson-CVPRW-2017.pdf}.

\bibitem[Agustsson et~al.(2019)Agustsson, Tschannen, Mentzer, Timofte, and Gool]{Agustsson_2019_ICCV}
Eirikur Agustsson, Michael Tschannen, Fabian Mentzer, Radu Timofte, and Luc~Van Gool.
\newblock Generative adversarial networks for extreme learned image compression.
\newblock In \emph{Proceedings of the IEEE/CVF International Conference on Computer Vision (ICCV)}, October 2019.

\bibitem[Agustsson et~al.(2023)Agustsson, Minnen, Toderici, and Mentzer]{agustsson2023multi}
Eirikur Agustsson, David Minnen, George Toderici, and Fabian Mentzer.
\newblock Multi-realism image compression with a conditional generator.
\newblock In \emph{Proceedings of the IEEE/CVF Conference on Computer Vision and Pattern Recognition}, pages 22324--22333, 2023.

\bibitem[Ball{\'e} et~al.(2017)Ball{\'e}, Laparra, and Simoncelli]{balle2017end}
Johannes Ball{\'e}, Valero Laparra, and Eero~P Simoncelli.
\newblock End-to-end optimized image compression.
\newblock In \emph{International Conference on Learning Representations}, 2017.

\bibitem[Ball{\'e} et~al.(2018)Ball{\'e}, Minnen, Singh, Hwang, and Johnston]{balle2018variational}
Johannes Ball{\'e}, David Minnen, Saurabh Singh, Sung~Jin Hwang, and Nick Johnston.
\newblock Variational image compression with a scale hyperprior.
\newblock \emph{6th International Conference on Learning Representations, {ICLR}}, 2018.

\bibitem[Bjontegaard(2001)]{bjontegaard2001calculation}
Gisle Bjontegaard.
\newblock Calculation of average psnr differences between rd-curves.
\newblock \emph{ITU SG16 Doc. VCEG-M33}, 2001.

\bibitem[Blau and Michaeli(2019)]{blau2019rethinking}
Yochai Blau and Tomer Michaeli.
\newblock Rethinking lossy compression: The rate-distortion-perception tradeoff.
\newblock In \emph{International Conference on Machine Learning}, pages 675--685. PMLR, 2019.

\bibitem[Bross et~al.(2021)Bross, Wang, Ye, Liu, Chen, Sullivan, and Ohm]{bross2021overview}
Benjamin Bross, Ye-Kui Wang, Yan Ye, Shan Liu, Jianle Chen, Gary~J Sullivan, and Jens-Rainer Ohm.
\newblock Overview of the versatile video coding (vvc) standard and its applications.
\newblock \emph{IEEE Transactions on Circuits and Systems for Video Technology}, 31\penalty0 (10):\penalty0 3736--3764, 2021.

\bibitem[Careil et~al.(2023)Careil, Muckley, Verbeek, and Lathuili{\`e}re]{careil2023towards}
Marlene Careil, Matthew~J Muckley, Jakob Verbeek, and St{\'e}phane Lathuili{\`e}re.
\newblock Towards image compression with perfect realism at ultra-low bitrates.
\newblock In \emph{The Twelfth International Conference on Learning Representations}, 2023.

\bibitem[Chang et~al.(2022)Chang, Zhang, Jiang, Liu, and Freeman]{chang2022maskgit}
Huiwen Chang, Han Zhang, Lu~Jiang, Ce~Liu, and William~T Freeman.
\newblock Maskgit: Masked generative image transformer.
\newblock In \emph{Proceedings of the IEEE/CVF conference on computer vision and pattern recognition}, pages 11315--11325, 2022.

\bibitem[Cheng et~al.(2020)Cheng, Sun, Takeuchi, and Katto]{cheng2020learned}
Zhengxue Cheng, Heming Sun, Masaru Takeuchi, and Jiro Katto.
\newblock Learned image compression with discretized gaussian mixture likelihoods and attention modules.
\newblock In \emph{Proceedings of the IEEE/CVF conference on computer vision and pattern recognition}, pages 7939--7948, 2020.

\bibitem[Dao et~al.(2024)Dao, Nguyen, Le, Vu, Nguyen, Pham, and Tran]{dao2024swiftbrush2}
Trung Dao, Thuan~Hoang Nguyen, Thanh Le, Duc Vu, Khoi Nguyen, Cuong Pham, and Anh Tran.
\newblock Swiftbrush v2: Make your one-step diffusion model better than its teacher.
\newblock In \emph{European Conference on Computer Vision}, pages 176--192. Springer, 2024.

\bibitem[Ding et~al.(2020)Ding, Ma, Wang, and Simoncelli]{ding2020image}
Keyan Ding, Kede Ma, Shiqi Wang, and Eero~P Simoncelli.
\newblock Image quality assessment: Unifying structure and texture similarity.
\newblock \emph{IEEE transactions on pattern analysis and machine intelligence}, 44\penalty0 (5):\penalty0 2567--2581, 2020.

\bibitem[Dong et~al.(2024)Dong, Fan, Guo, Wang, Zhang, Chen, Luo, and Zou]{dong2024tsd}
Linwei Dong, Qingnan Fan, Yihong Guo, Zhonghao Wang, Qi~Zhang, Jinwei Chen, Yawei Luo, and Changqing Zou.
\newblock Tsd-sr: One-step diffusion with target score distillation for real-world image super-resolution.
\newblock \emph{arXiv preprint arXiv:2411.18263}, 2024.

\bibitem[Esser et~al.(2021)Esser, Rombach, and Ommer]{esser2021taming}
Patrick Esser, Robin Rombach, and Bjorn Ommer.
\newblock Taming transformers for high-resolution image synthesis.
\newblock In \emph{Proceedings of the IEEE/CVF conference on computer vision and pattern recognition}, pages 12873--12883, 2021.

\bibitem[Franzen(1999)]{kodak}
Rich Franzen.
\newblock Kodak {PhotoCD} dataset, 1999.
\newblock URL \url{http://r0k.us/graphics/kodak/}.

\bibitem[Gokaslan et~al.(2023)Gokaslan, Cooper, Collins, Seguin, Jacobson, Patel, Frankle, Stephenson, and Kuleshov]{gokaslan2023commoncanvas}
Aaron Gokaslan, A~Feder Cooper, Jasmine Collins, Landan Seguin, Austin Jacobson, Mihir Patel, Jonathan Frankle, Cory Stephenson, and Volodymyr Kuleshov.
\newblock Commoncanvas: An open diffusion model trained with creative-commons images.
\newblock \emph{arXiv preprint arXiv:2310.16825}, 2023.

\bibitem[Guo et~al.(2025)Guo, Chen, Li, Guo, and Zhang]{guo2025compression}
Jinpei Guo, Zheng Chen, Wenbo Li, Yong Guo, and Yulun Zhang.
\newblock Compression-aware one-step diffusion model for jpeg artifact removal.
\newblock \emph{arXiv preprint arXiv:2502.09873}, 2025.

\bibitem[Heusel et~al.(2017)Heusel, Ramsauer, Unterthiner, Nessler, and Hochreiter]{heusel2017gans}
Martin Heusel, Hubert Ramsauer, Thomas Unterthiner, Bernhard Nessler, and Sepp Hochreiter.
\newblock Gans trained by a two time-scale update rule converge to a local nash equilibrium.
\newblock \emph{Advances in neural information processing systems}, 30, 2017.

\bibitem[Ho et~al.(2020)Ho, Jain, and Abbeel]{ho2020denoising}
Jonathan Ho, Ajay Jain, and Pieter Abbeel.
\newblock Denoising diffusion probabilistic models.
\newblock \emph{Advances in neural information processing systems}, 33, 2020.

\bibitem[Hu et~al.(2022)Hu, Shen, Wallis, Allen-Zhu, Li, Wang, Wang, Chen, et~al.]{hu2022lora}
Edward~J Hu, Yelong Shen, Phillip Wallis, Zeyuan Allen-Zhu, Yuanzhi Li, Shean Wang, Lu~Wang, Weizhu Chen, et~al.
\newblock Lora: Low-rank adaptation of large language models.
\newblock \emph{ICLR}, 1\penalty0 (2):\penalty0 3, 2022.

\bibitem[Jia et~al.(2024)Jia, Li, Li, Li, and Lu]{jia2024generative}
Zhaoyang Jia, Jiahao Li, Bin Li, Houqiang Li, and Yan Lu.
\newblock Generative latent coding for ultra-low bitrate image compression.
\newblock In \emph{Proceedings of the IEEE/CVF Conference on Computer Vision and Pattern Recognition}, pages 26088--26098, 2024.

\bibitem[Jiang et~al.(2023)Jiang, Yang, Zhai, Ning, Gao, and Wang]{jiang2023mlic}
Wei Jiang, Jiayu Yang, Yongqi Zhai, Peirong Ning, Feng Gao, and Ronggang Wang.
\newblock Mlic: Multi-reference entropy model for learned image compression.
\newblock In \emph{Proceedings of the 31st ACM International Conference on Multimedia}, pages 7618--7627, 2023.

\bibitem[Johnson et~al.(2016)Johnson, Alahi, and Fei-Fei]{johnson2016perceptual}
Justin Johnson, Alexandre Alahi, and Li~Fei-Fei.
\newblock Perceptual losses for real-time style transfer and super-resolution.
\newblock In \emph{Computer Vision--ECCV 2016: 14th European Conference, Amsterdam, The Netherlands, October 11-14, 2016, Proceedings, Part II 14}, pages 694--711. Springer, 2016.

\bibitem[Kang et~al.(2024)Kang, Zhang, Barnes, Paris, Kwak, Park, Shechtman, Zhu, and Park]{kang2024distilling}
Minguk Kang, Richard Zhang, Connelly Barnes, Sylvain Paris, Suha Kwak, Jaesik Park, Eli Shechtman, Jun-Yan Zhu, and Taesung Park.
\newblock Distilling diffusion models into conditional gans.
\newblock In \emph{European Conference on Computer Vision}, pages 428--447. Springer, 2024.

\bibitem[Kim and Kim(2024)]{kim2024tddsr}
Sohwi Kim and Tae-Kyun Kim.
\newblock Tddsr: Single-step diffusion with two discriminators for super resolution.
\newblock \emph{arXiv preprint arXiv:2410.07663}, 2024.

\bibitem[K{\"o}rber et~al.(2024{\natexlab{a}})K{\"o}rber, Kromer, Siebert, Hauke, Mueller-Gritschneder, and Schuller]{korber2024egic}
Nikolai K{\"o}rber, Eduard Kromer, Andreas Siebert, Sascha Hauke, Daniel Mueller-Gritschneder, and Bj{\"o}rn Schuller.
\newblock Egic: enhanced low-bit-rate generative image compression guided by semantic segmentation.
\newblock In \emph{European Conference on Computer Vision}, pages 202--220, 2024{\natexlab{a}}.

\bibitem[K{\"o}rber et~al.(2024{\natexlab{b}})K{\"o}rber, Kromer, Siebert, Hauke, Mueller-Gritschneder, and Schuller]{korber2024perco}
Nikolai K{\"o}rber, Eduard Kromer, Andreas Siebert, Sascha Hauke, Daniel Mueller-Gritschneder, and Bj{\"o}rn Schuller.
\newblock Perco ({SD}): Open perceptual compression.
\newblock In \emph{Workshop on Machine Learning and Compression, NeurIPS 2024}, 2024{\natexlab{b}}.
\newblock URL \url{https://openreview.net/forum?id=8xvygfdRWy}.

\bibitem[Lei et~al.(2023)Lei, Uslu, Hassani, and Bidokhti]{lei2023text+}
Eric Lei, Yi{\u{g}}it~Berkay Uslu, Hamed Hassani, and Shirin~Saeedi Bidokhti.
\newblock Text+ sketch: Image compression at ultra low rates.
\newblock \emph{arXiv preprint arXiv:2307.01944}, 2023.

\bibitem[Li et~al.(2023{\natexlab{a}})Li, Li, and Lu]{Li_2023_CVPR}
Jiahao Li, Bin Li, and Yan Lu.
\newblock Neural video compression with diverse contexts.
\newblock In \emph{Proceedings of the IEEE/CVF Conference on Computer Vision and Pattern Recognition (CVPR)}, pages 22616--22626, June 2023{\natexlab{a}}.

\bibitem[Li et~al.(2024{\natexlab{a}})Li, Li, and Lu]{li2024neural}
Jiahao Li, Bin Li, and Yan Lu.
\newblock Neural video compression with feature modulation.
\newblock In \emph{{IEEE/CVF} Conference on Computer Vision and Pattern Recognition, {CVPR} 2024, Seattle, WA, USA, June 17-21, 2024}, 2024{\natexlab{a}}.

\bibitem[Li et~al.(2023{\natexlab{b}})Li, Li, Savarese, and Hoi]{li2023blip}
Junnan Li, Dongxu Li, Silvio Savarese, and Steven Hoi.
\newblock Blip-2: Bootstrapping language-image pre-training with frozen image encoders and large language models.
\newblock In \emph{International conference on machine learning}, pages 19730--19742. PMLR, 2023{\natexlab{b}}.

\bibitem[Li et~al.(2024{\natexlab{b}})Li, Zhou, Wei, Ge, and Jiang]{li2024towards}
Zhiyuan Li, Yanhui Zhou, Hao Wei, Chenyang Ge, and Jingwen Jiang.
\newblock Towards extreme image compression with latent feature guidance and diffusion prior.
\newblock \emph{IEEE Transactions on Circuits and Systems for Video Technology}, 2024{\natexlab{b}}.

\bibitem[Li et~al.(2024{\natexlab{c}})Li, Zhou, Wei, Ge, and Mian]{li2024diffusion}
Zhiyuan Li, Yanhui Zhou, Hao Wei, Chenyang Ge, and Ajmal Mian.
\newblock Diffusion-based extreme image compression with compressed feature initialization.
\newblock \emph{arXiv preprint arXiv:2410.02640}, 2024{\natexlab{c}}.

\bibitem[Lin et~al.(2014)Lin, Maire, Belongie, Hays, Perona, Ramanan, Doll{\'a}r, and Zitnick]{lin2014microsoft}
Tsung-Yi Lin, Michael Maire, Serge Belongie, James Hays, Pietro Perona, Deva Ramanan, Piotr Doll{\'a}r, and C~Lawrence Zitnick.
\newblock Microsoft coco: Common objects in context.
\newblock In \emph{Computer Vision--ECCV 2014: 13th European Conference, Zurich, Switzerland, September 6-12, 2014, Proceedings, Part V 13}, pages 740--755. Springer, 2014.

\bibitem[Liu et~al.(2023)Liu, Sun, and Katto]{liu2023learned}
Jinming Liu, Heming Sun, and Jiro Katto.
\newblock Learned image compression with mixed transformer-cnn architectures.
\newblock In \emph{Proceedings of the IEEE/CVF conference on computer vision and pattern recognition}, pages 14388--14397, 2023.

\bibitem[Liu et~al.(2021)Liu, Lin, Cao, Hu, Wei, Zhang, Lin, and Guo]{liu2021swin}
Ze~Liu, Yutong Lin, Yue Cao, Han Hu, Yixuan Wei, Zheng Zhang, Stephen Lin, and Baining Guo.
\newblock Swin transformer: Hierarchical vision transformer using shifted windows.
\newblock In \emph{Proceedings of the IEEE/CVF international conference on computer vision}, pages 10012--10022, 2021.

\bibitem[Loshchilov and Hutter(2017)]{loshchilov2017decoupled}
Ilya Loshchilov and Frank Hutter.
\newblock Decoupled weight decay regularization.
\newblock \emph{arXiv preprint arXiv:1711.05101}, 2017.

\bibitem[Lu et~al.(2024)Lu, Xie, Jiang, Wang, Lin, and Wang]{lu2024hybridflow}
Lei Lu, Yanyue Xie, Wei Jiang, Wei Wang, Xue Lin, and Yanzhi Wang.
\newblock Hybridflow: Infusing continuity into masked codebook for extreme low-bitrate image compression.
\newblock In \emph{Proceedings of the 32nd ACM International Conference on Multimedia}, pages 3010--3018, 2024.

\bibitem[Mao et~al.(2024)Mao, Yang, Zhang, Wang, Wang, Wang, Jin, and Ma]{mao2024extreme}
Qi~Mao, Tinghan Yang, Yinuo Zhang, Zijian Wang, Meng Wang, Shiqi Wang, Libiao Jin, and Siwei Ma.
\newblock Extreme image compression using fine-tuned vqgans.
\newblock In \emph{2024 Data Compression Conference (DCC)}, pages 203--212, 2024.
\newblock \doi{10.1109/DCC58796.2024.00028}.

\bibitem[Mentzer et~al.(2020)Mentzer, Toderici, Tschannen, and Agustsson]{mentzer2020high}
Fabian Mentzer, George~D Toderici, Michael Tschannen, and Eirikur Agustsson.
\newblock High-fidelity generative image compression.
\newblock \emph{Advances in neural information processing systems}, 33:\penalty0 11913--11924, 2020.

\bibitem[Mentzer et~al.(2023)Mentzer, Minnen, Agustsson, and Tschannen]{mentzer2023finite}
Fabian Mentzer, David Minnen, Eirikur Agustsson, and Michael Tschannen.
\newblock Finite scalar quantization: Vq-vae made simple.
\newblock \emph{arXiv preprint arXiv:2309.15505}, 2023.

\bibitem[Minnen et~al.(2018)Minnen, Ball{\'e}, and Toderici]{minnen2018joint}
David Minnen, Johannes Ball{\'e}, and George~D Toderici.
\newblock Joint autoregressive and hierarchical priors for learned image compression.
\newblock \emph{Advances in neural information processing systems}, 31, 2018.

\bibitem[Muckley et~al.(2023)Muckley, El-Nouby, Ullrich, J{\'e}gou, and Verbeek]{muckley2023improving}
Matthew~J Muckley, Alaaeldin El-Nouby, Karen Ullrich, Herv{\'e} J{\'e}gou, and Jakob Verbeek.
\newblock Improving statistical fidelity for neural image compression with implicit local likelihood models.
\newblock In \emph{International Conference on Machine Learning}, pages 25426--25443. PMLR, 2023.

\bibitem[Nguyen and Tran(2024)]{nguyen2024swiftbrush}
Thuan~Hoang Nguyen and Anh Tran.
\newblock Swiftbrush: One-step text-to-image diffusion model with variational score distillation.
\newblock In \emph{Proceedings of the IEEE/CVF Conference on Computer Vision and Pattern Recognition}, pages 7807--7816, 2024.

\bibitem[Ohayon et~al.(2025)Ohayon, Manor, Michaeli, and Elad]{ohayon2025compressed}
Guy Ohayon, Hila Manor, Tomer Michaeli, and Michael Elad.
\newblock Compressed image generation with denoising diffusion codebook models.
\newblock \emph{arXiv preprint arXiv:2502.01189}, 2025.

\bibitem[OpenAI(2024)]{openai2024gpt4o}
OpenAI.
\newblock Gpt-4o.
\newblock \url{https://openai.com/index/gpt-4o}, 2024.
\newblock Accessed: 2025-05-05.

\bibitem[Qi et~al.(2025)Qi, Jia, Li, Li, Li, and Lu]{qi2025generative}
Linfeng Qi, Zhaoyang Jia, Jiahao Li, Bin Li, Houqiang Li, and Yan Lu.
\newblock Generative latent coding for ultra-low bitrate image and video compression.
\newblock \emph{IEEE Transactions on Circuits and Systems for Video Technology}, 2025.

\bibitem[Radford et~al.(2021)Radford, Kim, Hallacy, Ramesh, Goh, Agarwal, Sastry, Askell, Mishkin, Clark, et~al.]{radford2021learning}
Alec Radford, Jong~Wook Kim, Chris Hallacy, Aditya Ramesh, Gabriel Goh, Sandhini Agarwal, Girish Sastry, Amanda Askell, Pamela Mishkin, Jack Clark, et~al.
\newblock Learning transferable visual models from natural language supervision.
\newblock In \emph{International conference on machine learning}, pages 8748--8763. PmLR, 2021.

\bibitem[Rombach et~al.(2021)Rombach, Blattmann, Lorenz, Esser, and Ommer]{rombach2021highresolution}
Robin Rombach, Andreas Blattmann, Dominik Lorenz, Patrick Esser, and Björn Ommer.
\newblock High-resolution image synthesis with latent diffusion models, 2021.

\bibitem[Sohl-Dickstein et~al.(2015)Sohl-Dickstein, Weiss, Maheswaranathan, and Ganguli]{sohl2015deep}
Jascha Sohl-Dickstein, Eric Weiss, Niru Maheswaranathan, and Surya Ganguli.
\newblock Deep unsupervised learning using nonequilibrium thermodynamics.
\newblock In \emph{International conference on machine learning}, pages 2256--2265. pmlr, 2015.

\bibitem[Song et~al.(2020)Song, Meng, and Ermon]{song2020denoising}
Jiaming Song, Chenlin Meng, and Stefano Ermon.
\newblock Denoising diffusion implicit models.
\newblock \emph{arXiv preprint arXiv:2010.02502}, 2020.

\bibitem[Song et~al.(2023)Song, Dhariwal, Chen, and Sutskever]{song2023consistency}
Yang Song, Prafulla Dhariwal, Mark Chen, and Ilya Sutskever.
\newblock Consistency models.
\newblock In \emph{International Conference on Machine Learning}, pages 32211--32252. PMLR, 2023.

\bibitem[Song et~al.(2024)Song, Lorraine, Nie, Kreis, and Lucas]{song2024multi}
Yanke Song, Jonathan Lorraine, Weili Nie, Karsten Kreis, and James Lucas.
\newblock Multi-student diffusion distillation for better one-step generators.
\newblock \emph{arXiv preprint arXiv:2410.23274}, 2024.

\bibitem[Toderici et~al.(2020)Toderici, Shi, Timofte, Theis, Balle, Agustsson, Johnston, and Mentzer]{CLIC2020}
George Toderici, Wenzhe Shi, Radu Timofte, Lucas Theis, Johannes Balle, Eirikur Agustsson, Nick Johnston, and Fabian Mentzer.
\newblock Workshop and challenge on learned image compression (clic2020), 2020.
\newblock URL \url{http://www.compression.cc}.

\bibitem[von Platen et~al.(2022)von Platen, Patil, Lozhkov, Cuenca, Lambert, Rasul, Davaadorj, Nair, Paul, Berman, Xu, Liu, and Wolf]{diffusers}
Patrick von Platen, Suraj Patil, Anton Lozhkov, Pedro Cuenca, Nathan Lambert, Kashif Rasul, Mishig Davaadorj, Dhruv Nair, Sayak Paul, William Berman, Yiyi Xu, Steven Liu, and Thomas Wolf.
\newblock Diffusers: State-of-the-art diffusion models.
\newblock \url{https://github.com/huggingface/diffusers}, 2022.

\bibitem[Vonderfecht and Liu(2025)]{vonderfecht2025lossy}
Jeremy Vonderfecht and Feng Liu.
\newblock Lossy compression with pretrained diffusion models.
\newblock \emph{arXiv preprint arXiv:2501.09815}, 2025.

\bibitem[Wang et~al.(2023)Wang, Li, Li, and Lu]{wang2023EVC}
Guo-Hua Wang, Jiahao Li, Bin Li, and Yan Lu.
\newblock {EVC: Towards Real-Time Neural Image Compression with Mask Decay}.
\newblock In \emph{International Conference on Learning Representations}, 2023.

\bibitem[Wang et~al.(2024)Wang, Yang, Chen, Wang, Guo, Chau, Liu, Qiao, Kot, and Wen]{wang2024sinsr}
Yufei Wang, Wenhan Yang, Xinyuan Chen, Yaohui Wang, Lanqing Guo, Lap-Pui Chau, Ziwei Liu, Yu~Qiao, Alex~C Kot, and Bihan Wen.
\newblock Sinsr: diffusion-based image super-resolution in a single step.
\newblock In \emph{Proceedings of the IEEE/CVF conference on computer vision and pattern recognition}, pages 25796--25805, 2024.

\bibitem[Wang et~al.(2003)Wang, Simoncelli, and Bovik]{wang2003multiscale}
Z.~Wang, E.P. Simoncelli, and A.C. Bovik.
\newblock Multiscale structural similarity for image quality assessment.
\newblock In \emph{The Thrity-Seventh Asilomar Conference on Signals, Systems \& Computers, 2003}, volume~2, pages 1398--1402 Vol.2, 2003.
\newblock \doi{10.1109/ACSSC.2003.1292216}.

\bibitem[Wu et~al.(2024)Wu, Sun, Ma, and Zhang]{wu2024one}
Rongyuan Wu, Lingchen Sun, Zhiyuan Ma, and Lei Zhang.
\newblock One-step effective diffusion network for real-world image super-resolution.
\newblock \emph{Advances in Neural Information Processing Systems}, 37:\penalty0 92529--92553, 2024.

\bibitem[Xu et~al.(2025)Xu, Li, Li, Wang, Zhang, and Lu]{xu2025}
Tongda Xu, Jiahao Li, Bin Li, Yan Wang, Ya-Qin Zhang, and Yan Lu.
\newblock {PICD: Versatile Perceptual Image Compression with Diffusion Rendering}.
\newblock In \emph{{IEEE/CVF} Conference on Computer Vision and Pattern Recognition, {CVPR} 2025, Nashville, TN, USA, June 11-25, 2024}, 2025.

\bibitem[Xue et~al.(2024)Xue, Mao, Wang, Zhang, and Ma]{xue2024unifying}
Naifu Xue, Qi~Mao, Zijian Wang, Yuan Zhang, and Siwei Ma.
\newblock Unifying generation and compression: Ultra-low bitrate image coding via multi-stage transformer.
\newblock \emph{arXiv preprint arXiv:2403.03736}, 2024.

\bibitem[Xue et~al.(2025)Xue, Jia, Li, Li, Zhang, and Lu]{xue2025dlf}
Naifu Xue, Zhaoyang Jia, Jiahao Li, Bin Li, Yuan Zhang, and Yan Lu.
\newblock {DLF: Extreme Image Compression with Dual-generative Latent Fusion}.
\newblock In \emph{Proceedings of the IEEE/CVF International Conference on Computer Vision (ICCV)}, Oct 2025.

\bibitem[Yang et~al.(2023)Yang, Zhou, Feng, and Wang]{yang2023diffusion}
Xingyi Yang, Daquan Zhou, Jiashi Feng, and Xinchao Wang.
\newblock Diffusion probabilistic model made slim.
\newblock In \emph{Proceedings of the IEEE/CVF Conference on computer vision and pattern recognition}, pages 22552--22562, 2023.

\bibitem[Yin et~al.(2024{\natexlab{a}})Yin, Gharbi, Park, Zhang, Shechtman, Durand, and Freeman]{yin2024improved}
Tianwei Yin, Micha{\"e}l Gharbi, Taesung Park, Richard Zhang, Eli Shechtman, Fredo Durand, and Bill Freeman.
\newblock Improved distribution matching distillation for fast image synthesis.
\newblock \emph{Advances in neural information processing systems}, 37:\penalty0 47455--47487, 2024{\natexlab{a}}.

\bibitem[Yin et~al.(2024{\natexlab{b}})Yin, Gharbi, Zhang, Shechtman, Durand, Freeman, and Park]{yin2024one}
Tianwei Yin, Micha{\"e}l Gharbi, Richard Zhang, Eli Shechtman, Fredo Durand, William~T Freeman, and Taesung Park.
\newblock One-step diffusion with distribution matching distillation.
\newblock In \emph{Proceedings of the IEEE/CVF conference on computer vision and pattern recognition}, pages 6613--6623, 2024{\natexlab{b}}.

\bibitem[Yu et~al.(2023)Yu, Cheng, Wang, Kumar, Macherey, Huang, Ross, Essa, Bisk, Yang, et~al.]{yu2023spae}
Lijun Yu, Yong Cheng, Zhiruo Wang, Vivek Kumar, Wolfgang Macherey, Yanping Huang, David Ross, Irfan Essa, Yonatan Bisk, Ming-Hsuan Yang, et~al.
\newblock Spae: Semantic pyramid autoencoder for multimodal generation with frozen llms.
\newblock \emph{Advances in Neural Information Processing Systems}, 36:\penalty0 52692--52704, 2023.

\bibitem[Zhang et~al.(2023)Zhang, Rao, and Agrawala]{zhang2023adding}
Lvmin Zhang, Anyi Rao, and Maneesh Agrawala.
\newblock Adding conditional control to text-to-image diffusion models.
\newblock In \emph{Proceedings of the IEEE/CVF international conference on computer vision}, pages 3836--3847, 2023.

\bibitem[Zhang et~al.(2018)Zhang, Isola, Efros, Shechtman, and Wang]{zhang2018unreasonable}
Richard Zhang, Phillip Isola, Alexei~A Efros, Eli Shechtman, and Oliver Wang.
\newblock The unreasonable effectiveness of deep features as a perceptual metric.
\newblock In \emph{Proceedings of the IEEE conference on computer vision and pattern recognition}, pages 586--595, 2018.

\bibitem[Zhao et~al.(2023)Zhao, Li, Li, Xiong, and Lu]{zhao2023universal}
Jing Zhao, Bin Li, Jiahao Li, Ruiqin Xiong, and Yan Lu.
\newblock A universal optimization framework for learning-based image codec.
\newblock \emph{ACM Transactions on Multimedia Computing, Communications and Applications}, 20\penalty0 (1):\penalty0 1--19, 2023.

\bibitem[Zhou et~al.(2022)Zhou, Chan, Li, and Loy]{zhou2022towards}
Shangchen Zhou, Kelvin Chan, Chongyi Li, and Chen~Change Loy.
\newblock Towards robust blind face restoration with codebook lookup transformer.
\newblock \emph{Advances in Neural Information Processing Systems}, 35:\penalty0 30599--30611, 2022.

\end{thebibliography}
